\documentclass[journal,twocolumn]{IEEEtran}

\usepackage{amssymb}
\usepackage{amsthm,amsmath,amssymb}
\usepackage{mathrsfs}
\usepackage{amsfonts}
\usepackage{indentfirst}
\usepackage{graphicx}
\usepackage{subfigure}
\usepackage{float}
\usepackage{subeqnarray}
\usepackage{cases}
\usepackage{bm}
\usepackage{textcomp}
\usepackage{cite}
\usepackage{multirow}
\usepackage{enumerate}
\usepackage{color}

\makeatletter
\renewcommand{\maketag@@@}[1]{\hbox{\m@th\normalsize\normalfont#1}}%
\makeatother

\ifCLASSINFOpdf

\else

\fi

\makeatletter
\newif\if@restonecol
\makeatother

\usepackage[linesnumbered,ruled,vlined]{algorithm2e}
\usepackage{algpseudocode}

\newtheorem{theorem}{Theorem}
\newtheorem{lemma}{Lemma}
\newtheorem{assumption}{Assumption}
\newtheorem{definition}{Definition}
\newtheorem{corollary}{Corollary}

\begin{document}

\title{Analysis and Optimization of Wireless Federated Learning with Data Heterogeneity}

\author{Xuefeng Han, Jun Li, Wen Chen, Zhen Mei, Kang Wei, \\Ming Ding, and H. Vincent Poor
\thanks{Xuefeng Han and Wen Chen are with the Department of Electronic Engineering, Shanghai Jiao Tong University, Shanghai 200240, China (e-mail:
hansjell-watson@sjtu.edu.cn; wenchen@sjtu.edu.cn). Jun Li, Mei Zhen and Kang Wei are with School of Electrical and Optical
Engineering, Nanjing University of Science and Technology, Nanjing 210094,
China Ministry of Education (e-mail: \{jun.li; meizhen; kang.wei; \}@njust.edu.cn). Ming Ding is with Data61, CSIRO, Sydney, NSW 2015, Australia (e-mail:
ming.ding@data61.csiro.au). H. Vincent Poor is with the Department of Electrical and Computer Engineering, Princeton University, Princeton, NJ 08544 USA (e-mail: poor@princeton.edu).
}}

\markboth{}%
{Han \MakeLowercase{\textit{et al.}}: Application of Federated Learning in Wireless Communication}

\maketitle

\begin{abstract}
With the rapid proliferation of smart mobile devices, federated learning (FL) has been widely considered for application in wireless networks for distributed model training. \textcolor{black}{However, data heterogeneity, e.g., non-independently identically distributions and different sizes of training data among clients, poses major challenges to wireless FL. Limited communication resources complicate the implementation of fair scheduling which is required for training on heterogeneous data, and further deteriorate the overall performance. } To address this issue, this paper focuses on performance analysis and optimization for wireless FL, considering data heterogeneity, combined with wireless resource allocation. Specifically, we first develop a closed-form expression for an upper bound on the FL loss function, with a particular emphasis on data heterogeneity described by a dataset size vector and a data divergence vector. Then we formulate the loss function minimization problem, under constraints on long-term energy consumption and latency, and jointly optimize \underline{c}lient scheduling, \underline{r}esource allocation, and the number of local training \underline{e}pochs (CRE). Next, via the Lyapunov drift technique, we transform the CRE optimization problem into a series of tractable problems. Extensive experiments on real-world datasets demonstrate that the proposed algorithm outperforms other benchmarks in terms of the learning accuracy and energy consumption.
\end{abstract}

\IEEEpeerreviewmaketitle

\section{Introduction}
The modern era of artificial intelligence (AI) has witnessed powerful capabilities brought by machine learning (ML) in many applications, such as computer vision \cite{computer_vision}, autonomous vehicles \cite{autonomous_vehicles}, and so on. However, conventional ML algorithms require a centralized server to collect data from distributively located devices, consuming large amounts of communication resources. Also, privacy issues in the conventional ML are of growing concern, since data transmitted from the end devices may be eavesdropped upon during transmissions. To address these challenges, federated learning (FL), as a novel distributed learning paradigm, has been proposed~\cite{FedAvg}, in which a model is learned iteratively via local training on end devices and global aggregation on a server. Compared with centralized ML, FL has the advantages of reducing the communication burden by only transmitting models rather than raw data, balancing computational overload between the server and end devices for model training, and promoting privacy of clients by keeping data locally. In addition, FL can be incorporated with other advanced techniques, such as blockchain~\cite{blockchain, blockchain_and_allocation} and differential privacy~\cite{differential_privacy}, for enhancing its security.
\par
Meanwhile, with the development of Internet of Things (IoT) at unprecedented scales, mobile devices equipped with powerful hardware are capable of implementing ML algorithms for model training~\cite{edge_computing}. As a result, FL has been widely applied to various intelligent applications in wireless IoT networks. However, mobile devices usually have relatively limited computing and communication resources when executing FL tasks~\cite{book_ML, survey_ML}. First, limited computing resources at the mobile devices will cause inadequate local training, and thus degrade local model performance. Second, lack of communication resources, such as available channels and transmission power, may prevent devices from uploading their models successfully, therefore leading to an insufficient number of local models for global aggregation. Furthermore, data heterogeneity due to non-independently and non-identically distributed (non-IID) data and different data sizes of training samples on clients will result in significant divergence among local models, which is harmful for improving global model performance after aggregation.
\par
FL in wireless scenarios has drawn considerable attention recently. Much of the research in this area has concentrated on scheduling clients and allocating wireless resources to accelerate learning convergence and reduce the power consumption of the training process~\cite{FL_optimization_start, combined_metric, importance_of_local_updates, long_term_maximize_the_amount_of_data, reinforcement_learning_client_selection, importance-aware, local_epoch_and_round_with_model_fairness,  consumption_of_computation_and_communication, wireless_power_transfer}. The work in \cite{FL_optimization_start} designed an algorithm for selecting a proper number of local update epoches in each communication round. The work in \cite{combined_metric} defined a combined metric of each client which jointly considered the learning quality and the channel quality. To optimize the learning performance via client selecting, \cite{importance_of_local_updates, long_term_maximize_the_amount_of_data, reinforcement_learning_client_selection} designed different scheduling schemes under wireless constraints. Furthermore, constraints such as client fairness and computational capability have been taken into account in recent research on FL optimization. \textcolor{black}{The work in \cite{importance-aware} measured importance of samples among all clients and selected important samples to update the model.} The performance of wireless FL was enhanced in \cite{local_epoch_and_round_with_model_fairness} via weighted local loss functions with the consideration of model fairness. And \cite{consumption_of_computation_and_communication} derived the relationship between the performance and consumptions of computation and communication. Integrating wireless power transfer technology into FL, the work in \cite{wireless_power_transfer} constructed a tradeoff between model convergence and transmission power.
\par
\textcolor{black}{The preceding studies primarily concentrated on wireless resource allocation but often overlooked the issue of data heterogeneity among clients while optimizing performance under wireless constraints. In fact, achieving fair scheduling for training non-IID data contradicts the varying abilities of client participation due to differences in data sizes. Consequently, the above optimization methods are not suitable for addressing wireless FL with data heterogeneity.}
\par
\textcolor{black}{Some researchers attempted to handle this problem with additional datasets. For instance, in \cite{data_sharing_in_the_server}, the server shared a training dataset with clients to mitigate the highly non-IID characteristics of local datasets. In \cite{trust_score}, a dataset was used in the server to assign a trust score to each local model, which then influenced the updating of the global model based on the trust scores. Additionally, \cite{accuracy_compute} assessed the heterogeneity and fairness of local models by evaluating their performance on a common dataset. However, obtaining a suitable dataset for these methods in \cite{data_sharing_in_the_server, trust_score, accuracy_compute} was proved to be challenging, as it needs to guide models towards better performance. Moreover, using additional datasets adds to the computational burden, energy consumption, and latency. In contrast, \cite{asynchronous_federated_learning} proposed an asynchronous FL framework that employed a two-stage aggregation to mitigate the impact of non-IID data without relying on additional datasets. Nonetheless, this work overlooked the chaotic allocation of wireless resources in the asynchronous FL. \cite{distribution_acquire} scheduled clients based on the accurate distribution of all clients, and \cite{client_share} shared encoded local datasets among all clients to address data heterogeneity. However these requirements concerning local datasets are impractical in wireless FL, where data privacy is emphasized. Hence, applying the new frameworks of \cite{asynchronous_federated_learning, distribution_acquire, client_share} in wireless FL are challenging.}
\par
\textcolor{black}{Some other studies retained the basic FL framework while modifying training methods. As an example, \cite{heterogeneous_model} proposed different clients could train heterogeneous models based on their computation and communication capabilities. However, the comparisons to conventional optimization methods of wireless FL were insufficient, and the performance cost of training heterogeneous models lacked theoretical analysis. In \cite{federated_reinforcement_learning}, a divergence penalty term was introduced in the objective function to accelerate and stabilize the training process on heterogeneous data. Moreover, \cite{feature_shift} considered feature shift non-IID of sample inputs and utilized local batch normalization in the training process. Furthermore, \cite{lossy_compression_and_federated_dropout} employed a dropout method for local models, and \cite{different_epoch} normalized local models with different numbers of local epochs among clients. Nevertheless, these works did not
jointly optimize client scheduling and resource allocation in wireless networks. To sum up, the primary limitation of existing researches lies in the lack of a joint consideration of communication constraints and data heterogeneity for wireless FL.}
\par
In this paper, we are interested in designing an FL framework over wireless networks with a particular emphasis on data heterogeneity among clients. We aim to improve the performance of FL by jointly scheduling \underline{c}lients, allocating wireless \underline{r}esource and designing the number of local training \underline{e}pochs (CRE). To our best knowledge, this paper is the first work of its kind that attempts to investigate an adaptive FL scheme on non-IID data and unequal local dataset sizes under the constraints of wireless networks and client energy consumption. The main contributions can be summarized as follows.
\begin{itemize}
  \item We develop an upper bound on the loss function based on data heterogeneity expressed by a dataset size vector and a data divergence vector. Also, we design a mechanism to estimate the data divergence vectors and model property parameters utilized in the upper bound. Afterwards, a dynamic programming problem is formulated to minimize this bound under the constraints on latency and energy queue stability, via jointly optimizing client scheduling, transmission power, channel allocation, and the number of local epoches.
  \item We solve the optimization problem by transforming the long-term energy constraint into minimizing conditional Lyapunov drift. Then we formulate an equivalent optimization problem within each communication round and decompose it into two subproblems based on the Tammer decomposition method. The two subproblems are solved by alternating a closed-form solution of each single variable iteratively, and then employing the simulated annealing algorithm.
 \item Extensive experiments show that our proposed algorithm achieves the best performance under the strict latency and energy consumption constraints, compared to other benchmarks. \textcolor{black}{To be specific, our algorithm can improve the identification accuracy by up to 4.05\%, 4.13\%, 1.96\%, relative to the channel-allocate algorithm in \cite{importance_of_local_updates}, the importance-aware algorithm in \cite{importance-aware} and the FedNova algorithm in \cite{different_epoch}, respectively. Furthermore, our algorithm saves at least 46.34\% energy consumption compared to these algorithms.}
\end{itemize}
\par
The rest of this paper is organized as follows. Section II formulates the optimization problem. Section III develops the upper bound of the loss function of the FL framework. Section IV presents the solution to the optimization problem, and section VI draws conclusions. \textcolor{black}{The main notations in this paper are summarized in Table \ref{table: variable}.}
\section{System Model and Problem formulation}
\begin{figure*}[ht]
  \centering
  \includegraphics[width=0.75\textwidth]{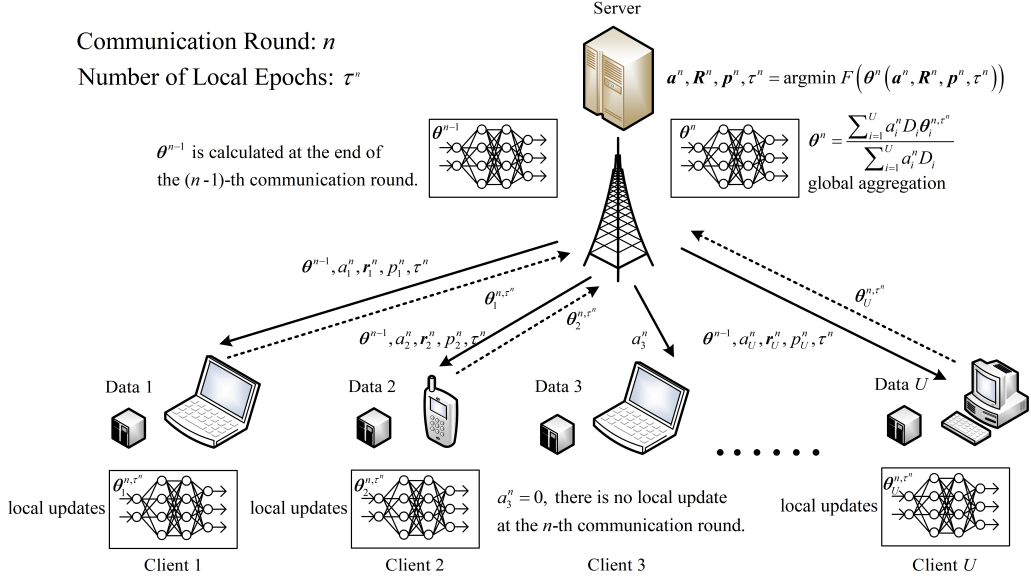}
  \caption{The architecture of FL process is over a wireless network consists of a server and many clients. Each participating client receives the broadcast, executes local updates and uploads its local model in parallel. If client $i$ receives $a_i^n=0$, it will not participate in the training process in the $n$-th communication round. The server waits a fixed latency, aggregates received local models and then starts a new communication round.}
  \label{figure: FL architecture}
\end{figure*}
Consider a wireless FL system consisting of $U$ clients and a centralized server for model aggregation. They cooperatively compute and communicate to accomplish a FL task by training a model $\bm \theta$ during $N$ communication rounds. Let  $ \mathcal{U} = \{1,2,\cdots, U\}$ and $\mathcal N = \{1, 2, \cdots, N\}$ denote the set of clients and the set of communication rounds, respectively. Client $i$, $i\in\mathcal U$ possesses a local dataset $\mathcal{D}_i$ and needs to fulfill local updates based on the local dataset. The set $\mathcal{D}_i$ can be expressed as $\mathcal{D}_i=\{ (\bm x_{l_i}, \bm y_{l_i})|l_i=1, 2, \cdots, D_i \}$, where $\bm x_{l_i}$ and $\bm y_{l_i}$ are the feature vector and the label vector of sample $l_i$, respectively, and $D_i$ is the size of $\mathcal{D}_i$. The server is co-located with the base station (BS) for performing global aggregations.
\par
Fig. \ref{figure: FL architecture} illustrates the procedures within the $n$-th communication round of a FL task. At the beginning, the server broadcasts the global model $\boldsymbol{\theta}^{n-1}$ to all clients. Client $i$ sets the global model as the initial local model by
$\boldsymbol{\theta}^{n, 0}_i := \boldsymbol{\theta}^{n-1}.$
Then, clients compute local loss functions and update local models in parallel. \textcolor{black}{The local loss function of client $i$ is
$F_i(\boldsymbol{\theta}_i^{n, m}) = \sum_{(\bm x_{l_i}, \bm y_{l_i}) \in \mathcal{D}_i} \frac{1}{D_i} f((\bm x_{l_i}, \bm y_{l_i}), \boldsymbol{\theta}_i^{n, m}),$ where $f((\bm x_{l_i}, \bm y_{l_i}), \boldsymbol{\theta}_i^{n, m})$ is the loss function of sample $(\bm x_{l_i}, \bm y_{l_i})$ on local model $\boldsymbol{\theta}_i^{n, m}$ at the $m$-th local epoch.} With the local gradient $\nabla F_i(\boldsymbol{\theta}^{n, m}_i)$, the local model at the $(m+1)$-th epoch is updated by
\begin{equation}
\boldsymbol{\theta}^{n, m+1}_i = \boldsymbol{\theta}^{n, m}_i - \eta^n \nabla F_i(\boldsymbol{\theta}^{n, m}_i),
\label{equation: local update}
\end{equation}
where $\eta^n$ is the learning rate. The number of local epochs for all clients in the $n$-th communication round is set into $\tau^n$. Since clients utilize batch gradient descent (BGD), the number of local updates is also $\tau^n$.
After $\tau^n$ epochs, client $i$ uploads the local model $\boldsymbol{\theta}^{n, \tau^n}_i$ to the server. In the server side, global aggregation is expressed as
\begin{equation}
\boldsymbol{\theta}^{n} = \sum^U_{i=1} \frac{D_i}{D} \boldsymbol{\theta}^{n, \tau^n}_{i} = \sum^U_{i=1} w_i \boldsymbol{\theta}^{n, \tau^n}_{i},
\label{equation: old global aggregation}
\end{equation}
where the aggregation weight $w_i$ is defined by $w_i = \frac{D_i}{D}$, and $D = \sum^U_{i=1} D_i$. It is noticed that the global aggregation requires synchronized local models, thus the server usually sets a maximal latency to avoid \textcolor{black}{waiting} too much. For $n=0$, the initial global model $\boldsymbol{\theta}^0$ is generated randomly by the server. Above local updates and global aggregation are performed in each communication round. A new communication round will start when the server broadcasts a new global model. \par
\subsection{Model Aggregation}
The limited communication resources result in only partial clients participating in a communication round. Therefore, we define $\bm{a}^n = [a^n_1, \cdots, a^n_U]^{\rm T}$ as the participation indicator vector, where $a^n_i \in \{0, 1\}$ denotes the participation state of client $i$ with $a^n_i = 1$ indicating that client $i$ participates in the $n$-th communication round. Denote by $\mathcal{U}_{\rm in}^n = \{i|a_i^n=1\}$ the set of participating clients and by $\mathcal{U}_{\rm out}^n = \{i|a_i^n=0\}$ the set of clients that do not participate. Thus, the global aggregation (\ref{equation: old global aggregation}) can be rewritten as
\begin{equation}
\boldsymbol{\theta}^{n} = \sum^{U}_{i=1} \frac{a^n_i D_i} {D^n} \boldsymbol{\theta}^{n, \tau^n}_i = \sum^U_{i=1} \tilde{w}_i^n \boldsymbol{\theta}^{n, \tau^n}_i,
\label{equation: new global aggregation}
\end{equation}
where the participation aggregation weight $\tilde{w}_i^n$ is defined as $\tilde{w}_i^n = \frac{a^n_i D_i} {D^n}$, and $D^n = \sum^U_{i=1} a^n_i D_i$. It is obvious that there is no contribution of client $i$ to the global model if $a_i^n=0$. Furthermore, we can see that (\ref{equation: new global aggregation}) degrades into (\ref{equation: old global aggregation}) when all clients participate. In the following, (\ref{equation: new global aggregation}) will be used to express the global aggregation in the $n$-th communication round.
\par
After $N$ communication rounds, the final model $\boldsymbol{\theta}^{N}$ is obtained, and the loss function of $\boldsymbol \theta^N$ on the entire dataset is expected to approach its minimum. Thus, the objective function of the global loss function $F(\bm \theta^N)$ can be written by
\begin{equation}
\min_{\boldsymbol{\theta}^N}\enspace F(\boldsymbol{\theta}^{N}) = \sum^U_{i=1} \frac{D_i}{D} F_i({\boldsymbol{\theta}^N})= \sum^U_{i=1}w_i F_i ({\boldsymbol{\theta}^N}).
\label{equation: initial aim}
\end{equation}
From (\ref{equation: new global aggregation}), it can be seen that $\boldsymbol{\theta}^n$ is closely related to $\bm{a}^n$. In order to minimize $F(\bm \theta^N)$ in (\ref{equation: initial aim}), it is vital to choose a series of participation vectors during the whole training process.
\begin{table*}[ht]
  \centering
  \caption{\textcolor{black}{List of Main Notations}}
\begin{tabular}{c||c|c||c}
\hline \hline
\textbf{Notation} & \textbf{Description} & \textbf{Notation} & \textbf{Description} \\ \hline
$U$ & Number of clients & $\mathcal U$ & Set of all clients \\ \hline
$C$ & Number of channels & $\mathcal U_{\rm in}^n$ & Set of participating clients in the $n$-th communication round \\ \hline
$D_i$ & Dataset size of client $i$ & $\mathcal U_{\rm out}^n$ & Set of absent clients in the $n$-th communication round \\ \hline
$\ell$ & Bit length of the model & $D^n$ & Sum of data sizes among participating clients in the $n$-th communication round \\ \hline
$a^n_i$ & Participation indicator of client $i$ & $w_i$ & Global aggregation weight of client $i$ with all client participating \\ \hline
$\boldsymbol a^n$ & Participation vector & $\tilde w_i^n$ & Participation aggregation weight of client $i$ in the $n$-th communication round \\ \hline
 $\boldsymbol R^n$ & Channel allocation matrix & $\boldsymbol \theta^n$ & Global model in the $n$-th communication round\\ \hline
 $\tau^n$ & Number of local epochs & $\boldsymbol \theta^{n,m}$ & Global model at the $m$-th epoch of the $n$-th communication round \\ \hline
 $t^{n, \rm{down}}$& Downlink latency & $\boldsymbol \theta^{n,m}_i$ & Local model of client $i$ at the $m$-th epoch of the $n$-th communication round \\ \hline
 $t_i^{n, \rm{up}}$ & Uplink latency of client $i$ & $\boldsymbol \phi^{n, m}$ & Virtual global model at $m$-th epoch of the $n$-th communication round \\ \hline
 $p_i^{n, \rm{up}}$ & Uplink power of client $i$ & $F(\boldsymbol \theta^n)$ & Global loss function of $\boldsymbol \theta^n$ with all client participating \\ \hline
 $\boldsymbol p^n$ & Uplink power vector & $\tilde F^n(\boldsymbol \theta^n)$ & Participation loss function of $\boldsymbol \theta^n$ in the $n$-th communication round \\ \hline
 $e_i^{n, \rm{up}}$ & Uplink energy consumption of client $i$ & $T_i$ & Computation latency for a local update of client $i$ \\ \hline
 $\eta^n$ & Learning rate & $E_i$ & Computation energy consumption for a local update of client $i$ \\ \hline
 $V$ & Lyapunov penalty factor & $p^{\max}$ & Maximal uplink power \\ \hline
 $\mathcal X^n$ & Solution of the optimal problem & $T^{\max}$ & Maximal latency for each communication round \\ \hline
 $d$ & Non-IID degree & $\sigma$ & Standard deviation of all dataset sizes \\
\hline \hline
\end{tabular}
\label{table: variable}
\end{table*}
\subsection{Model Transmissions}
\textcolor{black}{We now focus on the model transmission latency in both downlink and uplink periods. In the $n$-th communication round, the downlink utilizes the broadcast channel and its transmitting rate depends on the slowest one as
\begin{equation}
v^{n, {\rm down}} = \max_{i \in \mathcal U} \left\{ B^{\rm down} \log_2 \left( 1 + \frac{p^{{\rm down}} h^{n, {\rm down}}_i} {B^{\rm down} N_0} \right) \right\},
\label{equation: downlink rate}
\end{equation}
where $B^{\rm down}$ is the downlink bandwidth, $p^{{\rm down}}$ is the transmitting power of the server, $h_i^{n, {\rm down}}$ is the downlink channel gain to client $i$, and $N_0$ is the power spectrum density of noise. Note that the channel gain $h_{i}^{n, {\rm up}}$ contains large-scale fading $h_i^{n, {\rm large}}$, small-scale fading $h_{i}^{n, {\rm small}}$ and antenna gain $h^{\rm gain}$, i.e.,
$h^{n, {\rm up}}_{i} = h_i^{n, {\rm large}} h_{i}^{n, {\rm small}} h^{\rm gain}$. According to the urban macro (UMa) scenario described in \cite{channel_model}, the coefficient of large-scale fading can be expressed as $h_i^{n, {\rm large}} = 10^{- \frac{28 + 22 \lg \gamma_i + 20 \lg \nu}{10}}$, where $\gamma_i$ is the distance between the server and client $i$, $\nu$ is the carrier frequency. Considering frequency selectivity the multipath effect, $h_{i}^{n, {\rm small}}$ follows a $(K, \sigma)$ Rician distribution. After obtaining the transmitting rate, the downlink latency to transmit the previous global model is accordingly calculated by
\begin{equation}
t^{n, {\rm down}} = \frac{\ell(\boldsymbol{\theta}^{n-1})} {v^{n, {\rm down}}} = \frac{\ell}{v^{n, {\rm down}}},
\end{equation}}%
where $\ell(\boldsymbol{\theta}^{n-1})$ is the length of $\boldsymbol{\theta}^{n-1}$. Here, we assume the length of a model keeps constant during the whole training process. With $\boldsymbol \theta^{n-1}$ omitted, $\ell$ denotes the model length.
\par
As for the uplink transmission, each participating client uploads its local model via a pre-allocated private channel. We assume orthogonal frequency division multiple access (OFDMA) is adopted for uplink. Suppose that there are $C$ channels available and $\mathcal{C}$ denotes the set of all channels. Then, the allocation indicator vector is defined by  $\bm{r}_c^n=[r_{1,c}^n, r_{2,c}^n, \cdots , r_{U,c}^n]^{\rm T}$, $c\in \mathcal C$. The indicator $r_{i,c}^n\in \{0,1\}$ denotes the allocation state of channel $c$ to client $i$, where $r_{i,c}^n = 1$ indicates that channel $c$ is allocated to client $i$ in the $n$-th communication round, and $r^n_{i,c} = 0$, otherwise. Hence, the allocation constraint of channel $c$ yields
\begin{equation}
\sum^U_{i=1} r_{i,c}^n \leq 1.
\label{equation: channel constraint}
\end{equation}
For each participating client, it is essential to allocate a channel to upload its local model. Thus the allocation constraint for client $i$ is given by
\begin{equation}
\sum^C_{c=1} r_{i,c}^n = a_i^n.
\label{equation: access constraint}
\end{equation}
All allocation indicator vectors $\bm{r}^n_1, \bm{r}^n_2, \cdots, \bm{r}^n_C$ are stacked into a matrix $\bm{R}^n \in \{0, 1\}^{C \times U}$.
\par
Furthermore, the uplink transmission rate of client $i$ is calculated as
\begin{equation}
v_i^{n, {\rm up}} = \sum^C_{c=1} r^n_{i,c} B^{\rm up} \log_2 \left( 1 + \frac {p_i^{n, {\rm up}}h_{i, c}^{n, {\rm up}}} {B^{\rm up} N_0} \right),
\label{equation: uplink rate}
\end{equation}
where $B^{\rm up}$ is the bandwidth of each uplink channel, $p_i^{n, {\rm up}}$ is the transmitting power of client $i$, $h_{i, c}^{n, {\rm up}}$ is the channel gain of client $i$ for uplink channel $c$. \textcolor{black}{According to symmetry of downlink and uplink channels, $h_{i,c}^{n, {\rm up}}$ has a similar form to $h_i^{n, {\rm down}}$ and the only difference is that $h_i^{n, {\rm small}}$ is replaced by $h_{i, c}^{n, {\rm small}}$ due to OFDMA setting.} After obtaining the transmission rate of uplink in (\ref{equation: uplink rate}), the duration for model uploading can be expressed by
\begin{equation}
t^{n, {\rm up}}_i = \frac{\ell}{v^{n, {\rm up}}_i}.
\end{equation}
\par
Due to the limited energy of clients, we take into account the communication energy consumption for each client, i.e., $e_i^{n, {\rm up}} = p^{n, {\rm up}}_i t_i^{n, {\rm up}}$. Assume that devices of clients are same, and all clients have a same maximal transmission power constraint as
\begin{equation}
p^{n, {\rm up}}_i \leq p^{\max}.
\label{equation: power constraint}
\end{equation}
\par
Note that channel fading and transmitting power vary in different communication rounds, which directly affects the learning performance. Therefore, how to schedule clients and allocate channels properly is a key issue for learning performance.
\subsection{Model Local Training}
The computational resources are limited for each client. Therefore, it is essential to obtain the latency and energy consumption for local updates of each client. According to \cite{differential_privacy}, the computation latency of a local update is given by
$T_i = \frac{b D_i} {f},$
where $b$ is the number of CPU cycles computing a sample, and $f$ is the CPU frequency. We suppose that all clients are equipped with same computational capability. Based on \cite{computation_time}, the energy consumption of a local update for client $i$ is
$E_i = \alpha f^2 b D_i,$
where $\alpha$ is the energy consumption coefficient.
\par
It can be observed that the latency and energy consumption during local updates vary due to data heterogeneity. Thus, adapting computational resources is crucial to improve FL performance.
\subsection{Problem Formulation\label{subsection: optimal problem}}
Since local models are aggregated synchronously, the server sets a maximal waiting latency $T^{\max}$ for each communication round to receive local models from participating clients. Hence, a participating client needs to satisfy the latency constraint for uploading its local model as
\begin{equation}
\tau^n T_i + t^{n, {\rm down}} + t_i^{n, {\rm up}} \leq T^{\max},\quad i \in \mathcal U_{\rm in}^n,
\label{equation: latency constraint}
\end{equation}
The three terms in the left hand side of (\ref{equation: latency constraint}) are computational latency of executing $\tau^n$ local updates, downlink transmission latency and uplink transmission latency.
\par
Apart from the latency constraint, limited energy consumption at each client needs to be taken into account. The energy storage queue at client $i$ during the $n$-th communication round is represented by
\begin{equation}
q_i^{n} = E^{\rm add} - a_i^n(\tau^n E_i + e_i^{n, {\rm up}}),
\end{equation}
where $E^{\rm add}$ is the energy input to each client in each communication round, $\tau^n E_i$ is the energy consumption of executing $\tau^n$ local updates, and $e_i^{n,{\rm up}}$ is uplink communication energy. Since the energy consumption for computation and communication cannot exceed the storage, the energy constraint is given by
\begin{equation}
\sum_{k=1}^n q_i^k \geq 0, \quad n \in \mathcal N.
\label{equation: energy constraint}
\end{equation}
In the above analysis, $\tau^n$ acts an important role in both model updates in (\ref{equation: new global aggregation}) and constraints in (\ref{equation: latency constraint}) and (\ref{equation: energy constraint}). Taking these into account, we aim to design an adaptive number of local epochs. In addition, note that $\bm \theta^n$ is firmly related to $\tau^n$, and in the FL process, $\bm \theta^N$ is from a series of former models $\boldsymbol \theta^0, \bm \theta^1, \cdots, \bm \theta^{N-1}$. To reflect the model series, $F(\boldsymbol{\theta}^N)$ is subtracted by the initial loss function $F(\boldsymbol{\theta}^0)$, and the formula is split into
\begin{equation}
F(\boldsymbol{\theta}^N) - F(\boldsymbol{\theta}^0) = \sum^{N}_{n=1} [F(\boldsymbol{\theta}^n) - F(\boldsymbol{\theta}^{n-1})].
\label{equation: modified aim}
\end{equation}
\par
From constraint (\ref{equation: latency constraint}) and (\ref{equation: energy constraint}),  $\bm R^n$ and $\bm p^{n, {\rm up}}$ affect $\bm a^n$ and $\tau^n$, which are critical to $\bm \theta^n$. Hence, we can conclude that $\boldsymbol{\theta}^n$ is firmly related to client scheduling, channel allocation, uplink power controlling and the number of local epochs selection. To simplify the formula, $\textcolor{black}{\mathcal X^n} = \{\bm{a}^n, \bm{R}^n, \bm{p}^{n, {\rm up}}, \tau^n\}$ denotes the variable set which is solved in the $n$-th communication round. \textcolor{black}{$\mathcal X^n$} updates round by round to finally determine $\boldsymbol \theta^N$. Thus, the optimization problem is given by
\begin{align}
\textbf{P1:} \ & \min_{\{\textcolor{black}{\mathcal X^n}\}} \enspace \lim_{N \rightarrow \infty} \frac{1}{N}\sum^N_{n=1} [F( \boldsymbol{\theta}^n (\textcolor{black}{\mathcal X^n}, \boldsymbol{\theta}^{n-1}) ) - F(\boldsymbol{\theta}^{n-1})], \notag \\
{\rm s.t.} \textbf{C1:}\ & a_i^n,\thinspace r_{i,c}^n \in \{0, 1\}, \quad \forall i \in \mathcal{U}, \thinspace c \in \mathcal{C}, n \in \mathcal{N}, \notag\\
\textbf{C2:}\ & \sum^C_{c=1} r_{i,c}^n = a_i^n, \quad \forall i \in \mathcal{U}, n \in \mathcal{N}, \label{equation: optimal problem}\\
\textbf{C3:}\ & \sum^U_{i=1} r_{i,c}^n \leq 1, \quad \forall c \in \mathcal{C}, n \in \mathcal{N}, \notag\\
\textbf{C4:}\ & a_i^n p_i^{n, {\rm up}} \leq p^{\max}, \thinspace \forall i \in \mathcal{U}, n \in \mathcal N, \notag\\
\textbf{C5:}\ & \frac{1}{n}\sum^n_{k=1}q_i^k \geq 0, \quad  \forall i \in \mathcal{U}, n \in \mathcal N, \notag \\
\textbf{C6:}\ & a_i^n(\tau^n T_i + t^{n, {\rm down}} + t_i^{n, {\rm up}}) \leq T^{\max}, \forall i \in \mathcal{U}, n \in \mathcal{N}. \notag
\end{align}
In \textbf{P1}, $N \rightarrow \infty$ represents that the model has already converged, and \textcolor{black}{$\{ \mathcal X^n \} = \{ \mathcal X^0, \mathcal X^1, \cdots \}$}. However, it is not feasible to solve \textbf{P1} directly. First, the relation between $\bm \theta^n$ and $F(\bm \theta^n)$ is uncertain, and it is difficult to evaluate model $\boldsymbol \theta^n$ without testing on data. Second, the influence of \textcolor{black}{$\mathcal X ^n$} on $F(\boldsymbol \theta^n)$ needs to be analyzed. After all, \textcolor{black}{$\{ \mathcal X^n \}$} represents the solution that can minimize $\lim_{N \rightarrow \infty} \frac{1}{N}\sum^N_{n=1} [F( \boldsymbol{\theta}^n (\textcolor{black}{\mathcal X^n}, \boldsymbol{\theta}^{n-1}) ) - F(\boldsymbol{\theta}^{n-1})]$. Lastly, although $\bm \theta^{n - 1}$ and the channel state can be obtained in the $n$-th communication round, estimating them in the first communication round of the training process is difficult. Subsequently, we aim to derive a closed-form upper bound on the loss function, which can serve as the objective in the following optimization problem.
\section{Convergence Analysis\label{section: upper bound}}
In this section, a closed-form upper bound on the loss function is developed with a dataset size vector and a data divergence vector. The upper bound can consider data heterogeneity among clients with the two vectors. And for \textbf{P1}, generally speaking, $\boldsymbol{\theta}^n$ is a complicated neural network and there is no direct analytical relation between $\boldsymbol{\theta}^n$ and $\bm{X}^n$. However with some assumptions and definitions in \cite{FL_optimization_start} and \cite{FL_optimization_error}, some properties of $\bm \theta^n$ and $F(\bm \theta^n)$ can be obtained. Furthermore, numerical results in \cite{bound_tendency} indicate that an appropriate upper bound can capture the trend of the loss function.
\subsection{Definition and Assumption}
Some assumptions of the local loss function $F_i(\boldsymbol\theta)$ for $i \in \mathcal{U}$ are given as follows.
\begin{assumption}
Function $F_i(\boldsymbol{\theta})$ is convex.
\label{assumption: convex}
\end{assumption}
\begin{assumption}
Function $F_i(\boldsymbol{\theta})$ is $\rho$-Lipchitz, i.e., $| F_i(\boldsymbol{\theta}) - F_i(\boldsymbol{\theta}') | \leq \rho \| \boldsymbol{\theta} - \boldsymbol{\theta}' \|, \enspace \forall \boldsymbol\theta,\boldsymbol\theta'$.
\label{assumption: Lipschitz}
\end{assumption}
\begin{assumption}
Function $F_i(\boldsymbol\theta)$ is $\beta$-smooth, i.e., $\| \nabla F_i(\boldsymbol{\theta}) - \nabla F_i(\boldsymbol{\theta}') \| \leq \beta \| \boldsymbol{\theta} - \boldsymbol{\theta}' \|, \enspace \forall \boldsymbol\theta,\boldsymbol\theta'$.
\label{assumption: smooth}
\end{assumption}
\begin{assumption}
The distance between the gradient of the local loss function and the gradient of the global loss function has an upper bound $\delta_i$, i.e., $\| \nabla F_i(\boldsymbol\theta) - \nabla F(\boldsymbol\theta) \| \leq \delta_i$.
\label{assumption: divergence}
\end{assumption}
In above assumptions, $\rho$ and $\beta$ are defined as model property parameters. Also, the data divergence parameter $\delta_i$ is vital to reflect the non-IID feature of client $i$. However, $\rho, \beta$ and $\bm \delta = [\delta_1, \cdots, \delta_U]^{\rm T}$ are difficult to obtain in practical FL process. Hence at the start of each communication round, we iteratively estimate them with gradients and parameters of the model in the pervious communication round. Even if a client did not participate in the pervious communication round, its former gradients and parameters are still helpful. Therefore, based on \textbf{Assumption \ref{assumption: Lipschitz}}, \textbf{\ref{assumption: smooth}} and \textbf{\ref{assumption: divergence}}, parameters $\rho, \beta, \delta_i$ are estimated as
\begin{subequations}
\begin{align}
& \hat\rho^n = \max_{i \in \mathcal{U}_{\rm in}^{n-1}} \left\{\frac{ \left| F_i \left(\boldsymbol\theta_i^{n-1, \tau^{n-1}}\right) - F_i\left(\boldsymbol\theta^{n-1}\right) \right|}{ \left\| \boldsymbol\theta_i^{n-1, \tau^{n-1}} - \boldsymbol\theta^{n-1} \right\|} \right\},\\
& \hat\beta^n = \max_{i \in \mathcal{U}_{\rm in}^{n-1}} \left\{\frac{ \left\| \nabla F_i\left(\boldsymbol\theta_i^{n-1, \tau^{n-1}}\right) - \nabla F_i(\boldsymbol\theta^{n-1})\right\|}{\left\| \boldsymbol\theta_i^{n-1, \tau^{n-1}} - \boldsymbol\theta^{n-1} \right\|} \right\},\\
& \hat\delta^n_i = \max_{i,n} \left\{\left| \left\| \nabla F_i\left(\boldsymbol\theta^{n-1}\right) \right\| - \left\| \nabla F\left(\boldsymbol\theta^{n-1}\right) \right\| \right|\right\} \notag \\
& \qquad + \left\| \frac{\| \nabla F(\boldsymbol\theta^{n-1}) \|}{\| \nabla F_i(\boldsymbol\theta^{n-1}) \|} \nabla F_i\left(\boldsymbol\theta^{n-1}\right) - \nabla F\left(\boldsymbol\theta^{n-1}\right) \right\|,
\end{align}
\label{equation: parameter estimation}%
\end{subequations}
Note that the form of $\hat \delta_i^n$ in (\ref{equation: parameter estimation}c) is different from \textbf{Assumption 4}. If the difference between gradients is directly used on highly non-IID data, a positive feedback will occur and only a fixed partial clients will participate for all communication rounds. Hence, the local gradient is normalized and a bias is added in (\ref{equation: parameter estimation}c) to ensure that $\hat\delta_i^n$ satisfies \textbf{Assumption 4}.
\par
Guaranteeing estimated parameters in above assumptions, we continue to derive a lemma about the global loss function as follows.
\par
\begin{lemma}
Function $F(\boldsymbol{\theta})$ is convex, $\rho$-Lipschitz and $\beta$-smooth.
\label{lemma: global loss function property}
\end{lemma}
\begin{IEEEproof}
Note that $F(\bm \theta)$ is a weighted sum of $F_i(\bm \theta)$. With the assistance of the triangle inequality, \textbf{Lemma \ref{lemma: global loss function property}} can be proved according to \textbf{Assumption \ref{assumption: convex}-\ref{assumption: smooth}}.
\end{IEEEproof}
Although we have the above properties, it is tough to directly analyze the global loss function due to the varying partial client participation in different communication round. Hence we define another loss function and an auxiliary parameter as follows.
\begin{definition}
The loss function with the partial client participation $\tilde F^n(\boldsymbol\theta^n)$ is the weighted sum of local loss functions for $i \in \mathcal{U}_{\rm in}^n$, i.e.,
\begin{equation}
\tilde F^n(\boldsymbol\theta^n) \triangleq \sum_{i \in \mathcal{U}_{\rm in}^n} \frac{D_i}{D^n}F_i(\boldsymbol\theta^n) = \sum_{i \in \mathcal{U}_{\rm in}^n} \tilde w_i^n F_i(\boldsymbol\theta^n).
\label{equation: access loss function}
\end{equation}
\label{definition: access loss function}
\end{definition}
\begin{definition}
The auxiliary parameter vector $\boldsymbol\phi^{n, m}$ is initialized to $\boldsymbol\theta^n$ at the start of the $n$-th communication round and follows the centralized gradient descent with data of $\mathcal U_{\rm in}^n$. i.e.,
\begin{equation}
\boldsymbol\phi^{n,m} \triangleq \left\{
\begin{array}{ll}
\boldsymbol\theta^n, & \, m=0;\\
\boldsymbol\phi^{n, m-1} - \eta^n \nabla \tilde F^n(\boldsymbol\phi^{n, m-1}), & \, m = 1,\cdots, \tau^n.
\end{array}
\right.
\label{equation: auxiliary parameter vector}
\end{equation}
\label{definition: auxiliary parameter vector}
\end{definition}
\textcolor{black}{Different from $F(\bm \theta^n)$, $\tilde F^n(\bm \theta^n)$ only consists of participant clients rather than all clients. And $\bm \phi^{n,m}$ is updated by the centralized gradient descent rather than the distributed for $\bm \theta^{n,m}\triangleq \sum_{i \in \mathcal{U}^n_{\rm in}} \tilde w_i^n \bm \theta^{n,m}_i $, hence it can be served as a bridge between $\bm \theta^{n,m} $ and $\bm \theta_i^{n,m}$ in the later derivation.}
\subsection{Main Results\label{subsection: main results}}
\begin{theorem}
The upper bound on the difference between $\boldsymbol\theta^{n, m}$ and $\boldsymbol\phi^{n, m}$ is derived as
\begin{equation}
\| \boldsymbol\theta^{n, m} - \boldsymbol\phi^{n, m} \| \leq \frac{A_1}{\beta}((\eta^n \beta+1)^m - \eta^n \beta m - 1),
\end{equation}
where $A_1 = 2\sum^U_{i=1} (\tilde{w}_i^n - (\tilde{w}^n_i)^2) \delta_i$.
\label{theorem: parameter difference}
\end{theorem}
\begin{IEEEproof}
Please see Appendix \ref{subsection: proof of theorem 1}.
\end{IEEEproof}
\textbf{Theorem \ref{theorem: parameter difference}} gives an upper bound of the difference between models trained distributedly and centralizedly. We can observe that when there is only one participating client or $\tau^n$ is set into 0, the upper bound will vanish to 0, which is consistent with the realistic situation. After analyzing $\bm \phi^{n, m}$ with \textbf{Theorem \ref{theorem: parameter difference}}, the upper bound of $F(\bm \phi^{n, m})$ is given by \textbf{Theorem \ref{theorem: loss of auxiliary parameter}}.
\begin{theorem}
When $\eta^n \beta < 1$ is satisfied, $F(\boldsymbol\phi^{n, m})$ is bounded by
\begin{equation}
\begin{aligned}
& F(\boldsymbol\phi^{n, m}) - F(\boldsymbol\theta^*) \\
& \leq \frac{2m A_3}{-1 + \sqrt{1 + \frac{4mA_3} {F(\boldsymbol\phi^{n, 0}) - F(\boldsymbol\theta^*)} + \frac{(4\eta^n - 2 (\eta^n)^2 \beta)m^2 A_3} {B_1^2}}} ,
\end{aligned}
\label{equation: loss of auxiliary parameter}
\end{equation}
where $A_3 = (\eta^n - (\eta^n)^2\beta) \sqrt{2\beta A_2 (F(\boldsymbol{\theta}^{n,0}) - F(\boldsymbol{\theta}^*))} + \frac{(\eta^n)^2 \beta A_2}{2}$, $A_2 = 2 (1- \sum^U_{j=1} a_j^n w_j^n) \times \sum^U_{i=1} (\tilde{w}_i^n + w_i^n - 2 a_i^n w_i^n) \delta_i^2$,  $B_1 = \max_{n, m} \|\bm{\phi}^{n, m} - \boldsymbol{\theta}^*\|$.
\label{theorem: loss of auxiliary parameter}
\end{theorem}
\begin{IEEEproof}
Please see Appendix \ref{subsection: proof of theorem 2}.
\end{IEEEproof}
In \textbf{Theorem \ref{theorem: loss of auxiliary parameter}}, we provide the upper bound of $F(\bm \phi^{n, m})$ whose model is centralizedly trained on a part of clients. Due to its complex form in (\ref{equation: loss of auxiliary parameter}), only a specific situation is analyzed: when all clients participate, $A_3$ will vanish to 0 and this bound will attain its minimum. Following the former analysis, the upper bound of the $F(\bm \theta^{n})$ is obtained by \textbf{Corollary \ref{corollary: loss upper bound}}.
\begin{corollary}
The upper bound of $F(\boldsymbol\theta^{n})$ at the end of the $n$-th communication round is
\begin{equation}
\begin{aligned}
F(\boldsymbol{\theta}^{n}) - F(\boldsymbol{\theta}^*) \leq & \frac{\rho A_1} {\beta} ((\eta^n \beta + 1)^{\tau^n} - \eta^n \beta \tau^n -1) + \tau^n A_3\\
& + \frac{2}{\frac{(2\eta^n - (\eta^n)^2\beta)}{B_1^2}\tau^n + \frac{2}{F(\boldsymbol{\theta}^{n, 0}) - F(\boldsymbol{\theta}^*)}}.
\end{aligned}
\label{equation: loss upper bound}
\end{equation}
\label{corollary: loss upper bound}
\end{corollary}
\begin{IEEEproof}
With \textbf{Theorem \ref{theorem: parameter difference}} and \textbf{Lemma \ref{lemma: global loss function property}}, the difference of global loss functions has a upper bound, that is, $| F(\boldsymbol\theta^{n, m}) - F(\boldsymbol\phi^{n, m}) | \leq \frac{\rho A_1}{\beta}((\eta^n \beta+1)^m - \eta^n \beta m - 1)$. Then, replacing $m$ by $\tau^n$ in \textbf{Theorem \ref{theorem: loss of auxiliary parameter}} and subtracting both sides by $F(\boldsymbol \theta^*)$, the inequality is transformed into
\begin{equation}
\begin{aligned}
& F(\boldsymbol\phi^{n, \tau^n}) - F(\boldsymbol\theta^*) \\
& \leq \frac{2\tau^n A_3}{-1 + \sqrt{1 + \frac{4\tau^n A_3} {F(\boldsymbol\phi^{n, 0}) - F(\boldsymbol\theta^*)} + \frac{(4\eta^n - 2 (\eta^n)^2 \beta)(\tau^n)^2 A_3} {B_1^2}}}.
\end{aligned}
\end{equation}
To simplify the radical expression, formulas for the difference of square and Taylor's first-order approximation are used to obtain
\begin{equation}
\begin{aligned}
& F(\boldsymbol\phi^{n, \tau^n}) - F(\boldsymbol\theta^*) \\
& \leq \tau^n A_3 + \frac{2}{\frac{(2\eta^n - (\eta^n)^2\beta)}{B_1^2}\tau^n + \frac{2}{F(\boldsymbol{\phi}^{n, 0}) - F(\boldsymbol{\theta}^*)}}.
\end{aligned}
\end{equation}
With $\boldsymbol \phi^{n, 0} = \boldsymbol \theta^{n-1}$, $\boldsymbol \theta^{n, \tau^n} = \boldsymbol\theta^n$ and $|F(\boldsymbol\theta^{n, \tau^n}) - F(\boldsymbol\theta^*)| \leq | F(\boldsymbol\theta^{n, \tau^n}) - F(\boldsymbol\phi^{n, \tau^n}) | + | F(\boldsymbol\phi^{n, \tau^n}) - F(\boldsymbol\theta^*) |$, we can derive an upper bound of $|F(\boldsymbol\theta^{n, \tau^n}) - F(\boldsymbol\theta^*)|$. Since $\boldsymbol\theta^*$ is optimal, the absolute value sign of $| F(\boldsymbol\theta^{n, \tau^n}) - F(\boldsymbol\theta^*) |$ can be omitted. Hence, we have proved \textbf{Corollary \ref{corollary: loss upper bound}}.
\end{IEEEproof}
\textcolor{black}{In \textbf{Corollary \ref{corollary: loss upper bound}}, $\tau^n$ and $\boldsymbol a^n$ exactly determine the upper bound value. Computing the derivate of the upper bound with respect to $\tau^n$, we can conclude that the upper bound firstly decreases and then increases. Such a conclusion is consistent to the fact that too small $\tau^n$ leads to slow convergence and too large $\tau^n$ induces divergent local models and an unstable global model. As for $\boldsymbol a^n$, it is noticed that clients with small $\delta_i$ could conduce to a small bound according to the form of $A_1$ and $A_3$. Furthermore, $A_3$ also encourages more participating clients. Above analytical formulas and conclusions vitally contribute to the following optimization.}
\section{Problem Solution\label{section: solution}}
In this section, we propose a method to solve \textbf{P1}. With the assistance of the Lyapunov optimization method in \cite{Lyapunov_energy} and the upper bound in Section \ref{section: upper bound}, the intractable optimization problem \textbf{P1} is first transformed to a deterministic optimization problem \textbf{P2} for each communication round. And then, \textbf{P2} can be decomposed into two subproblems which are both solved individually.
\subsection{Problem Transformation}
Based on the Lyapunov optimization framework in \cite{Lyapunov_optimization}, the time average inequality of the energy can be transformed into a queue stability constraint. Primarily, a virtual queue $Z^n_i$ is defined by $Z^n_i = \max\{Z^{n-1}_i - q^{n-1}_i, 0\}$. Thus, constraint \textbf{C5} is transformed equivalently into
$\lim_{N \rightarrow \infty} \frac{|Z_i^N|}{N}=0$. The perturbed Lyapunov function is defined by $L^n = \frac{1}{2} \sum_{i=1}^{U} (Z_i^n)^2$. Then, the expected conditional Lyapunov drift is $\Delta^n = \mathbb{E} \{L^{n+1} - L^n | Z_i^n \}$. Enlarging $\Delta^n$ by $(Z_i^{n+1})^2 \leq (Z_i^{n} - q^{n}_i)^2$, and the conditional Lyapunov drift is bounded by
\begin{equation}
\Delta^n \leq \mathbb{E} \left\{ \left. \sum^U_{i=1} \left((q_i^n)^2 - 2 Z_i^n q_i^n\right) \right| Z_i^n \right\}.
\label{equation: conditional Lyapunov drift upper bound}
\end{equation}
Adding the penalty about the objective function and the Lyapunov drift-plus-penalty function is
\begin{equation}
\begin{aligned}
\Delta_V^n = \Delta^n + V \mathbb{E} & \left\{ F\left( \boldsymbol{\theta}^n (\{\bm{a}^n, \bm{R}^n, \bm{p}^{n,\rm{up}}, \tau^n\}, \boldsymbol{\theta}^{n-1}) \right) \right. \\
& \enspace \left.- F(\boldsymbol{\theta}^{n-1}) \left. \right| Z_i^n \right\},
\end{aligned}
\label{equation: Lyapunov drift-plus-penalty function}
\end{equation}
where $V \geq 0$ is the Lyapunov penalty factor to tune the trade-off between the descent of the loss function and the energy queue stability. Large $V$ can emphasize the performance of the model, and vice versa. Especially, $V \rightarrow \infty$ means only maximizing the descent regardless of energy consumption. On the other hand, when $V = 0$, only the stability of the energy queue is ensured. We remark that the expectation in (\ref{equation: Lyapunov drift-plus-penalty function}) is related to the uncertainty of current channel states and the former model. When CREs work in the $n$-th communication round, the uncertainty of channel states can be eliminated by the channel estimation \cite{channel_estimation} and $\bm \theta^{n-1}$ can be obtained, thus the expectation sign is removed. Then, substituting (\ref{equation: loss upper bound}) and (\ref{equation: conditional Lyapunov drift upper bound}), the upper bound of (\ref{equation: Lyapunov drift-plus-penalty function}) is
\begin{equation}
\begin{aligned}
& J(\bm{a}^n, \bm{R}^n, \bm{p}^{n,\rm{up}}, \tau^n) - V (F(\boldsymbol \theta^{n-1}) - F(\boldsymbol \theta^*))\\
& = \sum^U_{i = 1} [(q_i^n)^2 - 2 Z_i^n q_i^n]  - V (F(\boldsymbol \theta^{n-1}) - F(\boldsymbol \theta^*)) \\
& \quad + \frac{\rho V A_1} {\beta} ((\eta^n \beta + 1)^{\tau^n} - \eta^n \beta \tau^n -1) + \tau^n A_3 V \\
& \quad + \frac{2V}{\frac{(2\eta^n - (\eta^n)^2\beta)}{B_1^2}\tau^n + \frac{2}{F(\boldsymbol{\theta}^{n, 0}) - F(\boldsymbol{\theta}^*)}}.
\end{aligned}
\label{equation: objective function}
\end{equation}
Note that the term $ V (F(\boldsymbol \theta^{n-1}) - F(\boldsymbol \theta^*))$ can be omitted since it keeps constant in the $n$-th communication round, thus the original problem in \textbf{P1} is rewritten as
\begin{align}
& \textbf{P2:} \quad \min_{\bm{a}^n, \bm{R}^n, \bm{p}^{n,\rm{up}}, \tau^n} \quad J(\bm{a}^n, \bm{R}^n, \bm{p}^{n,\rm{up}}, \tau^n), \notag \\
{\rm s.t.} \quad & \textbf{C4}'\textbf{:} \quad p_i^{n, {\rm up}} \leq p^{\max}, \quad  \forall i \in \mathcal{U}, n \in \mathcal{N}, \label{equation: modified optimal problem} \\
& \textbf{C1}, \textbf{C2}, \textbf{C3}, \textbf{C6}, \notag
\end{align}
where $a_i^n$ is omitted in $\textbf{C4}'$, on account of $p_i^{n, \rm{up}}$ for $ i \in \mathcal{U}_{\rm{out}}^n$ having no concern with \textbf{P2}. Now, the long-term problem has been transformed into (\ref{equation: modified optimal problem}) in each communication round. At the beginning of the $n$-th communication round, $(\bm{a}^n, \bm{R}^n, \bm{p}^{n,\rm{up}}, \tau^n)$ is solved by (\ref{equation: modified optimal problem}), and then, the server broadcasts communication command and $\bm \theta^{n-1}$ to clients. After downlink communication, local updates, uplink communication and the global aggregation, $\boldsymbol{\theta}^n$ is obtained and the $n$-th communication round is over. Finally, new channel responses and model property parameters are estimated in the server. Hence, another new optimization problem in the $(n+1)$-th communication round is proposed in the same way and so on.
\par
It can be seen that the problem in (\ref{equation: modified optimal problem}) is a mixed integer nonlinear program (MINLP). Since it is too complex to solve (\ref{equation: modified optimal problem}) directly, a low-complexity solution will be designed next.
\subsection{Problem Decomposition}
Consider categories of variables $(\bm{a}^n, \bm{R}^n, \bm{p}^{n,\rm{up}}, \tau^n)$: $\bm{a}^n$ and $\bm{R}^n$ are combinatorial variables, $\bm{p}^{n, \rm{up}}$ is the continuous variable, and $\tau^n$ is the integral variable. To simplify the problem, $\tau^n$ can go slack into the continuous variable. Motivated by \cite{heuristic_method}, Tammer decomposition method is employed to transform (\ref{equation: modified optimal problem}) into an equivalent master problem with an inner subproblem and an outer subproblem. The master problem is written as
\begin{equation}
\begin{aligned}
& \textbf{P3:} \min_{\bm{a}^n, \bm{R}^n} \left( \min_{\tau^n, \bm{p}^{n, \rm{up}}} J(\bm{a}^n, \bm{R}^n, \bm{p}^{n,\rm{up}}, \tau^n) \right),\\
{\rm s.t.} \enspace & \textbf{C1}, \textbf{C2}, \textbf{C3}, \textbf{C4}', \textbf{C6}.
\end{aligned}
\label{equation: master problem}
\end{equation}
In (\ref{equation: master problem}), the outer problem is about $\bm{a}^n$ and $\bm{R}^n$, and \textbf{C1}, \textbf{C2}, \textbf{C3}, \textbf{C5} are constraints with respect to this problem. Hence, the outer problem \textbf{P3.1} is given by
\begin{equation}
\textbf{P3.1:} \min_{\bm{a}^n, \bm{R}^n} \ J_1(\bm{a}^n, \bm{R}^n), \quad \textbf{s.t.} \enspace \textbf{C1}, \textbf{C2}, \textbf{C3}, \textbf{C6}
\label{equation: outer problem}
\end{equation}
where $J_1(\bm{a}^n, \bm{R}^n) = J(\bm a^n, \bm R^n, \bm p^{n, {\rm up}*}, \tau^{n*})$ is the optimal value of the inner problem \textbf{P3.2}, i.e.,
\begin{equation}
\textbf{P3.2:} \min_{\tau^n, \bm{p}^{n, \rm{up}}} J(\bm{a}^n, \bm{R}^n, \bm{p}^{n,\rm{up}}, \tau^n), \quad \textbf{s.t.} \enspace \textbf{C4}', \textbf{C6},
\label{equation: inner problem}
\end{equation}
where $\textbf{C4}'$ and \textbf{C6} are constraints of $\tau^n, \bm{p}^{n, \rm{up}}$.
\subsection{Continuous Optimization \label{subsection: continuous optimization}}
Deleting constant terms in (\ref{equation: objective function}) with fixed $\bm{a}^n,\bm{R}^n$, we have
\begin{equation}
\begin{aligned}
& J_2(\tau^n, \bm{p}^{n, {\rm up}}) \\
& = \sum_{i \in \mathcal{U}_{\rm in}^n} (Z_i^n + \tau^n E_i + e_i^n - E^{\rm add})^2 \\
& \quad + \sum_{i \in \mathcal{U}_{\rm out}^n}\left[(E^{\rm add})^2 -2Z_i^n E^{\rm add} \right] + \tau^n A_3 V \\
& \quad + \frac{\rho A_1 V}{\beta}((\eta^n\beta+1)^{\tau^n} - \eta^n \beta \tau^n - 1)  \\
& \quad + \frac{2V}{\frac{(2\eta^n - (\eta^n)^2\beta)}{B_1^2}\tau^n + \frac{2}{F(\boldsymbol{\theta}^{n-1}) - F(\boldsymbol{\theta}^*)}}.
\end{aligned}
\label{equation: developed bound}
\end{equation}
Hence, the inner continuous optimization problem \textbf{P3.2} is converted into
\begin{equation}
\textbf{P3.2}'\textbf{:}\min_{\tau^n, \bm{p}^{n, {\rm up}}} J_2(\tau^n, \bm{p}^{n, {\rm up}}), \quad
\textbf{s.t.} \enspace \textbf{C4}', \textbf{C6}.
\end{equation}
\par
Note that $J_2(\tau^n, \bm{p}^{n, {\rm up}})$ is convex with respect to $\tau^n$. This is due to the fact that the first partial derivative and the second partial derivative of $J_2(\tau^n, \bm{p}^{n, {\rm up}})$ with respect to $\tau^n$ can be calculated from (\ref{equation: developed bound}) and we can prove that $\frac{\partial^2 J_2} {\partial (\tau^n)^2}$ is positive.
\par
Another variable need to be analyzed is the uplink power $\bm{p}^{n, {\rm up}}$. For client $i$ in $\mathcal{U}_{\rm out}^n$, $p_i^{n, {\rm up}}$ is zero. Hence, only uplink powers of participating clients are optimized. Although the first partial derivative and the second partial derivative of $J_2(\tau^n, \bm{p}^{n, {\rm up}})$ with respect to $p^{n, {\rm up}}_i$ can be calculated from (\ref{equation: developed bound}), we find the sign of $\frac{\partial^2 J_2}{(\partial p_i^{n, {\rm up}})^2}$ is uncertain. Nevertheless, it is easy to prove the first derivative $\frac{d e_i^{n, {\rm up}}}{d (p_i^{n, {\rm up}})}>0$, and the sign of $\frac{\partial J_2}{\partial p_i^{n, {\rm up}}}$ only depends on $e_i^{n, {\rm up}} + Z_i^n + \tau^n E_i - E^{\rm add}$. Hence, the monotonicity of $J_2$ with respect to $p_i^{n, {\rm up}}$ can be obtained.
\par
It is easy to optimize $\bm p^{n, {\rm up}}$ or $\tau^n$, however, it is difficult to optimize them jointly. To solve this problem, \cite{alternating_iteration_continuous} adopted an alternating iterative method, which is a common way to solve the continuous optimization problem. In this way, we fix $\bm{p}^{n, {\rm up}}$ and the problem about $\tau^n$ is
\begin{equation}
\begin{aligned}
& \textbf{P3.2}'\textbf{.1:} \min_{\tau^n} \enspace J_2(\tau^n, \bm{p}^{n, {\rm up}}), \\
{\rm s.t.} \enspace & \textbf{C6:} \enspace \tau^n T_i + t^{n, \rm down} + t_i^{n, \rm up} \leq T^{\max}, \quad \forall i \in \mathcal{U}_{\rm in}^n.
\end{aligned}
\label{equation: optimization tau}
\end{equation}
In \textbf{C6}, the maximum number of local epochs can be calculated by
\begin{equation}
\begin{aligned}
& \tau^{n, \max} = \\
& \min_{i \in \mathcal{U}_{\rm in}^n} \left\{\frac{1}{T_i} \left[ T^{\max} - t^{n, \rm down} - \frac{\ell}{B^{\rm up} \log_2(1 + \frac{p_i^{n, \rm up}h_{i, c}^n}{B^{\rm up}N_0}) } \right] \right\}.
\end{aligned}
\label{equation: epoch maximum}
\end{equation}
Then, on account of the convexity of $J_2$ with respect to $\tau^n$, the monotonicity of $J_2(\tau^n)$ can be determined. The first derivative at two special points (0 and $+\infty$) are $\frac{\partial J_2}{\partial \tau^n}|_{\tau^n=0} = \sum_{i \in \mathcal{U}_{\rm in}^n}[2E_i(Z_i^n + e_i^{n, \rm up} - E^{\rm add})] + \frac{\rho A_1 V}{\beta}( \ln(\eta^n \beta + 1) - \eta^n \beta) + A_3 V - \frac{V(2\eta^n - (\eta^n)^2 \beta)(F(\boldsymbol{\theta}^{n,0}) - F(\boldsymbol{\theta}^*))^2}{2B_1^2}$ and $\frac{\partial J_2}{\partial \tau^n}|_{\tau^n \rightarrow +\infty} = 0$. If $\frac{\partial J_2}{\partial \tau^n}|_{\tau^n=0} < 0$, and there must be $\tau'$ satisfying $\frac{\partial J_2}{\partial \tau^n}|_{\tau^n = \tau'} = 0$. After computing $\frac{\partial J_2}{\partial \tau^n}$ and comparing $\tau'$ and $\tau^{n, \max}$, the optimal number of local epochs is
\begin{equation}
\tau^{n*} = \left\{
\begin{array}{ll}
1, \enspace & \frac{\partial J_2}{\partial \tau^n}|_{\tau^n=0} \geq 0; \\
\tau' , \enspace & \frac{\partial J_2}{\partial \tau^n}|_{\tau^n=0} < 0 \enspace {\rm and} \enspace \tau' \leq \lfloor \tau^{n, \max} \rfloor; \\
\lfloor \tau^{n, \max} \rfloor, \enspace & \frac{\partial J_2}{\partial \tau^n}|_{\tau^n=0} < 0 \enspace {\rm and} \enspace \tau' > \lfloor \tau^{n, \max} \rfloor.
\end{array}
\label{equation: optimal epoch}
\right.
\end{equation}
\par
For the uplink power, the optimization problem is
\begin{align}
& \textbf{P3.2}'\textbf{.2:} \quad \min_{\bm{p}^{n, \rm up}} \enspace J_2(\tau^n, \bm{p}^{n, \rm up}), \label{equation: optimization power} \\
\rm{s.t.} \ & \textbf{C7:} \enspace p_i^{n, \rm up} = 0, \quad \forall i \in \mathcal{U}_{\rm out}^n, \notag \\
& \textbf{C8:} \enspace p_i^{n, \rm up} \leq p^{\max}, \quad \forall i \in \mathcal{U}_{\rm in}^n, \notag \\
& \textbf{C9:} \enspace \tau^n T_i + t^{n, \rm down} + \frac{\ell}{B^{\rm up} \log_2(1+\frac{p^{n, \rm up}_i h_{i,c}^n}{B^{\rm up}N_0})} \leq T^{\max}, \notag \\
& \qquad \forall i \in \mathcal{U}_{\rm in}^n. \notag
\end{align}
In \textbf{C9}, the uplink power of participating clients is larger than
$ p_i^{n, \rm up, \min} = (2^{\frac{\ell}{B^{\rm up} (T^{\max} - t^{n, \rm down} - \tau^n T_i)}}-1) \frac{B^{\rm up}N_0}{h_{i,c}^n}$.
The minimum and the maximum of $e_i^{n, \rm up}$ are expressed by
$e_i^{n, \rm up, \max} = \frac{p^{\max} \ell}{B^{\rm up} \log_2(1+\frac{p^{\rm \max} h_{i,c}^n}{B^{\rm up}N_0})}$ and $e_i^{n, \rm up, min} = (p_i^{n, \rm up, \min}) (T^{\max} - t^{n, \rm down} - \tau^n T_i)$, respectively.
Moreover, due to $\frac{d e_i^{n, \rm up}}{d (p_i^{n, \rm up})}>0$, the monotonicity of $J_2(\bm{p}^{n, \rm up})$ only depends on $e_i^{n, \rm up} + Z_i^n + \tau^n E_i - E^{\rm add}$. Based on the relation between $e_i^{n, \rm up}$ and $p_i^{n, \rm up}$, with $e_i^{\rm up'}=E^{\rm add} - Z_i^n-\tau^n E_i$, the zero point of $\frac{\partial J_2}{\partial (p_i^{n, \rm up})}$ is
\begin{equation}
\begin{aligned}
& p_i^{\rm up'} \\
& = -\frac{e_i^{n, \rm up}B^{\rm up} W_{-1}(-\frac{\ell N_0 \ln 2}{e_i^{n, \rm up}h_{i, c}^n} \exp(-\frac{\ell N_0 \ln 2}{e_i^{n, \rm up}h_{i, c}^n}))}{\ell \ln 2} - \frac{B^{\rm up}N_0}{h_{i, c}^n},
\end{aligned}
\label{equation: power zero point}
\end{equation}
where $W_{-1}(\cdot)$ is the lower part of the Lambert W function. By comparing the feasible region and the zero point, the optimal uplink power of participating clients is given by
\begin{equation}
p_i^{n, \rm up*} = \left\{
\begin{array}{ll}
p_i^{n, \rm up, \min}, & \quad e_i^{\rm up'} < e_i^{n, \rm up, \min}; \\
p_i^{\rm up'}, & \quad  e_i^{n, \rm up, \min} \leq e_i^{\rm up'} \leq e_i^{n, \rm up, \max}; \\
p^{\max}, & \quad e_i^{\rm up'} > e_i^{n, \rm up, \max}.
\end{array}
\right.
\label{equation: optimal power}
\end{equation}
\par
Overall, a full description of our optimization method is presented in \textbf{Algorithm \ref{algorithm: number and power}}.
\begin{algorithm}
  \caption{Joint Number of Local Epochs Design and Uplink Power Allocation Algorithm}
  \label{algorithm: number and power}
  \KwIn{participation state vector $\bm{a}^n$ and channel allocation matrix $\bm{R}^n$}
  \KwOut{optimal number of local epochs $\tau^{n*}$ and optimal uplink power vector $\bm{p}^{n, {\rm up}*}$}
  Set $s = 1$, $\bm{p}^0 = \bm{0}$, $\tau^0=1$ and set convergence threshold $\epsilon_p$, $\epsilon_\tau$\;
  Initialize $p^1_i$ for $i \in \mathcal{U}_{\rm in}^n$ randomly, set $p^1_i=0$ for $i \in \mathcal{U}_{\rm out}^n$ and calculate $\tau^1 = \lfloor \tau^{n, \max} \rfloor$ according to (\ref{equation: epoch maximum})\;
  \While{$\|\bm p^s - \bm p^{s-1}\| > \epsilon_p$ or $|\tau^s - \tau^{s-1}| > \epsilon_\tau$}
  {
    Update $s:=s+1$\;
    Update $\tau^s$ according to (\ref{equation: optimal epoch}) with $\bm{p}^{s-1}$\;
    Update $\bm{p}^s$ according to (\ref{equation: optimal power}) with $\tau^s$\;
  }
  Return $\tau^{n*} = \lfloor \tau^s \rfloor$ and $\bm p^{n, \rm up*}$ according to (\ref{equation: optimal power}) with $\lfloor \tau^s \rfloor$ .
\end{algorithm}
\subsection{Combinatorial Optimization \label{subsection: combinatorial optimization}}
To reduce the complexity of the combinatorial optimization problem \textbf{P3.1}, the simulated annealing algorithm is adopted. Since $\bm a^n$ and $\bm R^n$ should satisfy \textbf{C2} in \textbf{P3.1}, we only set $\bm R^n$ as the variable and express $\bm a^n$ with $\bm R^n$. To enable the heuristic search of the simulated annealing algorithm, the neighboring set of $\bm R^n$ is defined as follows.
\begin{definition}
Each 0-1 matrix $\bm{R}^{s}_j$ in the neighboring set $\mathcal{R}^s=\{\bm{R}_1^s, \cdots, \bm{R}_{|\mathcal{R}^s|}^s \}$ satisfies \textbf{C3} and that its Euclid distance to $\bm{R}^{s-1}$ is no more than 1. i.e, the neighboring set is defined by
\begin{equation}
\begin{aligned}
& \mathcal{R}^s = \\
& \left\{ \bm R_j^s \in \{0, 1\}^{C \times U} \Big | \sum^U_{i=1} (r_{i,c})_j \leq 1, \| \bm{R}_j^s - \bm{R}^{s-1} \| \leq 1 \right\}.
\end{aligned}
\label{equation: neighboring set}
\end{equation}
\label{definition: neighboring set}
\end{definition}
In the beginning, channel allocation matrix is zeroed as $\bm R^{s}$ and $s=0$. Then, randomly select a matrix from the neighboring set $\mathcal R^s$ of $\bm R^{s-1}$ as $\bm R^{s}_j$. According to the pervious introduction, $\bm a_j^s, \tau_j^s, \bm p_j^s$ can be obtained. Next, the new objective function $J^s_j$ is compared with the the former result $J^{s - 1}$ and receive $(\bm a^s_j, \bm R^s_j, \tau^s_j, \bm p^s_j)$ with a probability. Then, execute the next iteration until the maximal iteration number $s_{\max}$. The detailed process is presented in \textbf{Algorithm \ref{algorithm: channel and access}}.
\begin{algorithm}
  \caption{Simulated Annealing Algorithm}
  \label{algorithm: channel and access}
  \KwOut{optimal channel allocation matrix $\bm{R}^{n*}$, optimal access state vector $\bm{a}^{n*}$, optimal local epoch number $\tau^{n*}$, optimal uplink power vector $\bm p^{n*}$}
  Set annealing temperature $T$, decay rate $\alpha$, maximal iteration number $s_{\max}$\;
  Initialize $\bm{R}^0 = \bm{0}$, $\bm a^0 = \bm 0$, $\tau^0 = 0$, $\bm p^0 = 0$, $J^0 = +\infty$ and set $s = 0$\;
  \While{$s < s_{\max}$}
  {
    Update $s:= s+1$ and update the neighboring set $\mathcal{R}^s$ of $\bm R^{s-1}$ according to (\ref{equation: neighboring set})\;
    Choose a matrix $\bm R_j^s$ randomly from $\mathcal{R}^s$ and update $\bm a_j^s$ with $\bm R_j^s$ according to (\ref{equation: modified optimal problem}c)\;
    Compute the optimal local epoch number $\tau^s_j$ and the optimal uplink power vector $\bm p^s_j$ with $(\bm a_j^s, \bm R_j^s)$ according to \textbf{Algorithm \ref{algorithm: number and power}}\;
    Update the objective function $J_j^s$ with $(\bm a_j^s, \bm R_j^s, \tau^s_j, \bm p^s_j)$ according to (\ref{equation: objective function})\;
    \eIf{$J_j^s < J^{s-1}$}
    {
        Update $(\bm a^s, \bm R^s, \tau^s, \bm p^s) := (\bm a_j^s, \bm R_j^s, \tau^s_j, \bm p^s_j)$ and $J^s := J^s_j$\;
    }
    {
        Update $(\bm a^s, \bm R^s, \tau^s, \bm p^s) := (\bm a_j^s, \bm R_j^s, \tau^s_j, \bm p^s_j)$ and $J^s := J^s_j$ with $Pr = e^{\frac{J_j^s - J^{s-1}}{T}}$, otherwise, update $(\bm a^s, \bm R^s, \tau^s, \bm p^s) := (\bm a^{s-1}, \bm R^{s-1}, \tau^{s-1}, \bm p^{s-1})$ and $J^s := J^{s-1}$\;
    }
    Update the annealing temperature $T = \alpha T$\;
  }
  Search the minimal objective function $J^*$, and return $(\bm a^{n*}, \bm R^{n*}, \tau^{n*}, \bm p^{n*})$.
\end{algorithm}
\textcolor{black}{
\subsection{Computational Complexity and Convergence Analysis of the Proposed Algorithm}
\subsubsection{Computational Complexity} The complexity of alternating iterations relies on prescribed precision parameters, and the total number of iterations is in the order of $\log(\frac{1}{\epsilon_p\epsilon_\tau})$. Furthermore, there are at most $C$ optimization problems of uplink powers and 1 optimization problem of epoch numbers within each iteration in \textbf{Algorithm \ref{algorithm: number and power}}. In \textbf{Algorithm \ref{algorithm: channel and access}}, the complexity is related to the maximal iteration number $s_{\max}$. Hence, the complexity of proposed algorithms is $O\left(Cs_{\max} \log(\frac{1}{\epsilon_\tau\epsilon_p})\right)$.
\subsubsection{Convergence Analysis} During the inner continuous optimization process, $\bm p^s$ and $\bm \tau^s$ are solved alternately. Due to the optimality of each variable, we have $J_2(\tau^s, \bm p^{s-1}) \leq J_2(\tau^{s-1}, \bm p^{s-1})$ and $J_2(\tau^s, \bm p^s) \leq J_2(\tau^s, \bm p^{s-1})$ for each iteration, which educes $J_2(\tau^r, \bm p^r) \leq J_2(\tau^{s-1}, \bm p^{s-1})$ .Since $J_2$ is non-increasing and has a infinite lower bound, the convergence of \textbf{Algorithm \ref{algorithm: number and power}} can be guaranteed. And in \textbf{Algorithm \ref{algorithm: channel and access}}, the probability of accepting a worse solution tends to 0 as $T$ decays, which guarantees the convergence of the combinatorial optimization process. Hence, the convergence of proposed algorithms is proved.
}
\section{Simulation Results\label{section: result}}
In our simulations, a circular network area consisting of a BS and 10 clients within a radius of $500 \rm m $ is considered. \textcolor{black}{All clients are uniformly distributed in the circular area. As for training data, each client possesses a unique dataset with various data sizes and a non-IID data setting is adopted in our simulations.} Other parameters used are listed in Table~\ref{table: parameter value}. Computational devices of clients are same and channel responses and datasets are different. \par
\begin{table}[ht]
  \centering
  \caption{System Parameters}
  \begin{tabular}{c|c|c|c}
    \hline \hline
    \textbf{Parameter} & \textbf{Value} & \textbf{Parameter} & \textbf{Value} \\ \hline
    $p^{\rm down}$ & 1 W & $p^{\rm max}$ & 0.2 W \\ \hline
    $B^{\rm down}$ & 20 MHz & $B^{\rm up}$ & 1 MHz \\ \hline
    $K_{\rm Rician}$ & 4 & $\sigma_{\rm Rician}$ & 1 \\ \hline
    $h^{\rm gain}$ & 65 dB & $N_0$ & -174 dBm/Hz \\ \hline
    $\alpha$ & $10^{-26}$ & $b$ & 100 \\ \hline
    $f$ & $5 \times 10^8$ & $E^{\rm add}$ & 0.00175 J \\ \hline
    $t_{\rm MLP, max}$ & 0.01 s & $t_{\rm CNN, max}$ & 0.1 s \\ \hline
    $\ell_{\rm MLP}$ & 318080 bits & $\ell_{\rm CNN}$ & 3781200 bits\\ \hline
    \hline
  \end{tabular}
  \label{table: parameter value}
\end{table}\par
\begin{figure*}[ht]
  \centering
  \subfigure[Accuracy on MNIST ]{\includegraphics[width=0.38\textwidth]{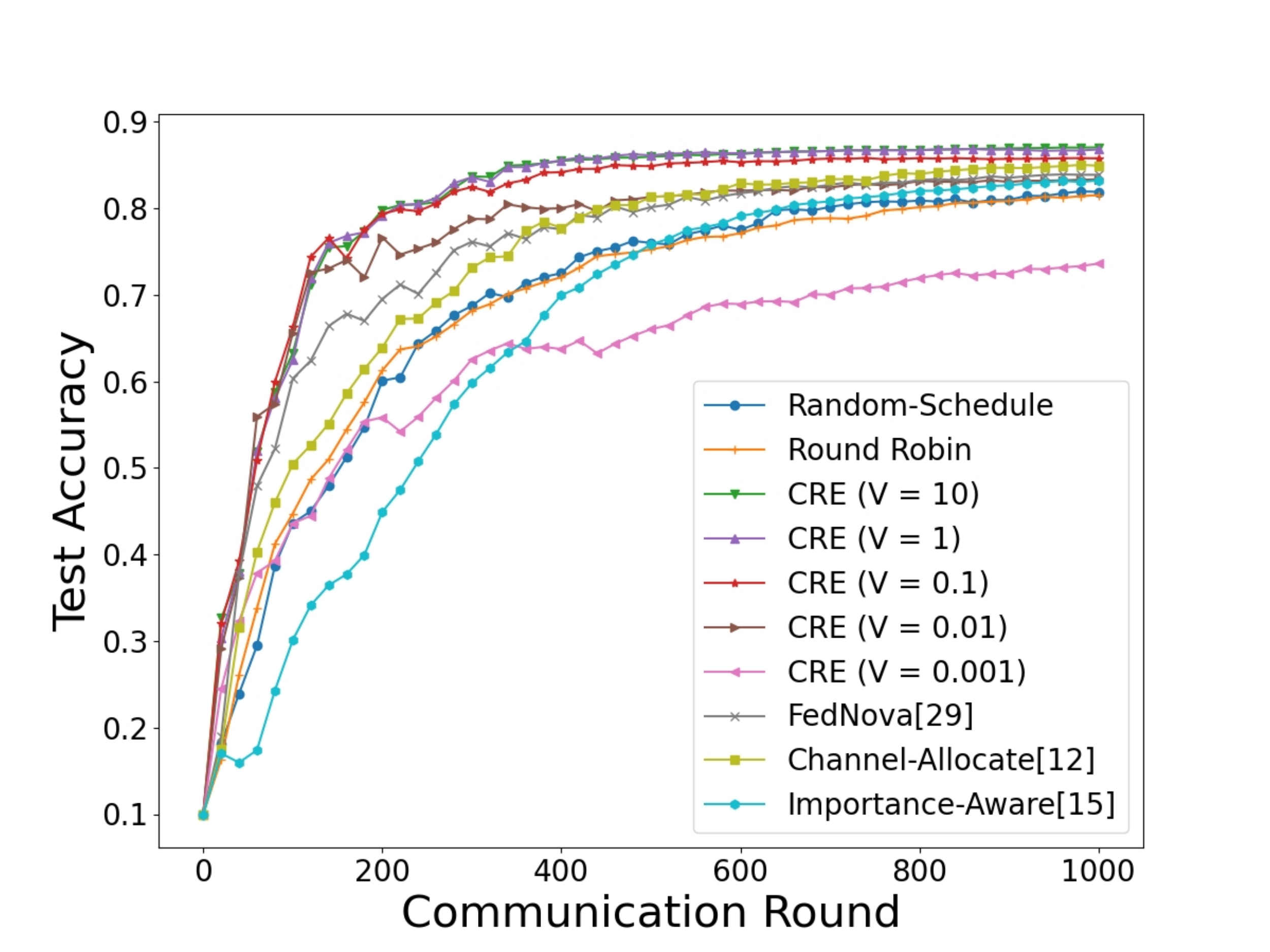}}
  \subfigure[Energy on MNIST
  ]{\includegraphics[width=0.38\textwidth]{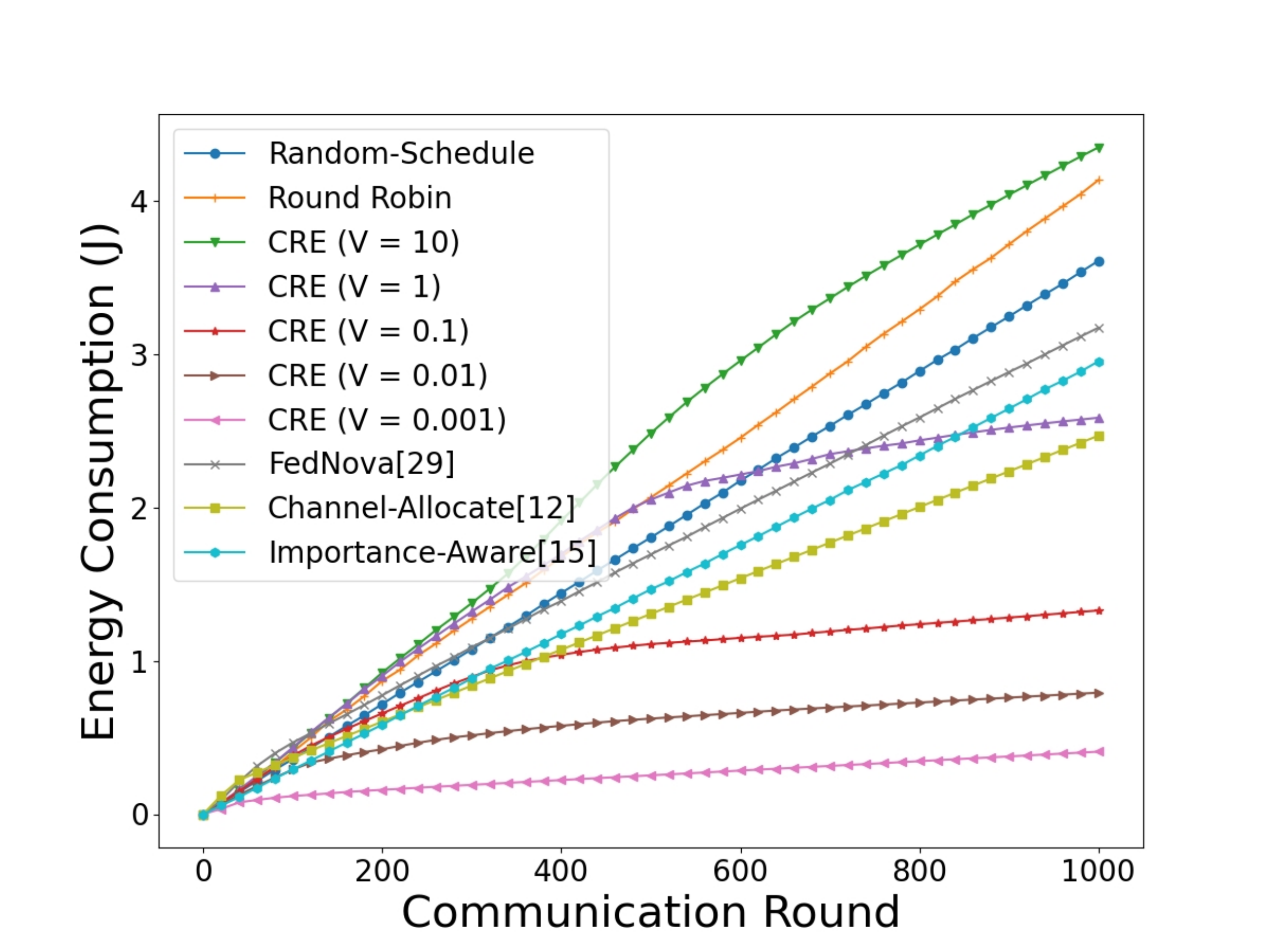}}
  \\
  \subfigure[Accuracy on CIFAR-10
  ]{\includegraphics[width=0.38\textwidth]{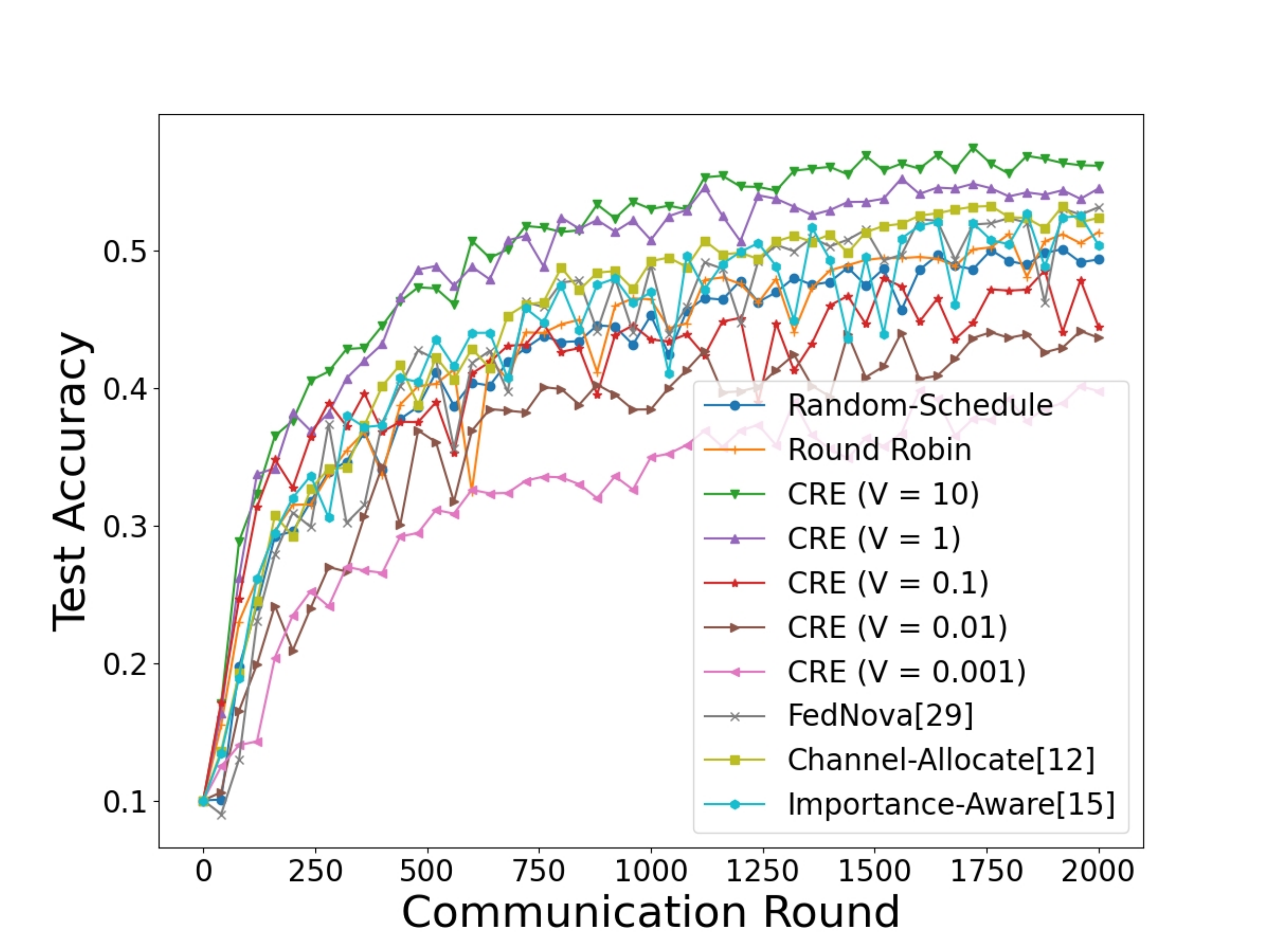}}
  \subfigure[Energy on CIFAR-10 ]{\includegraphics[width=0.38\textwidth]{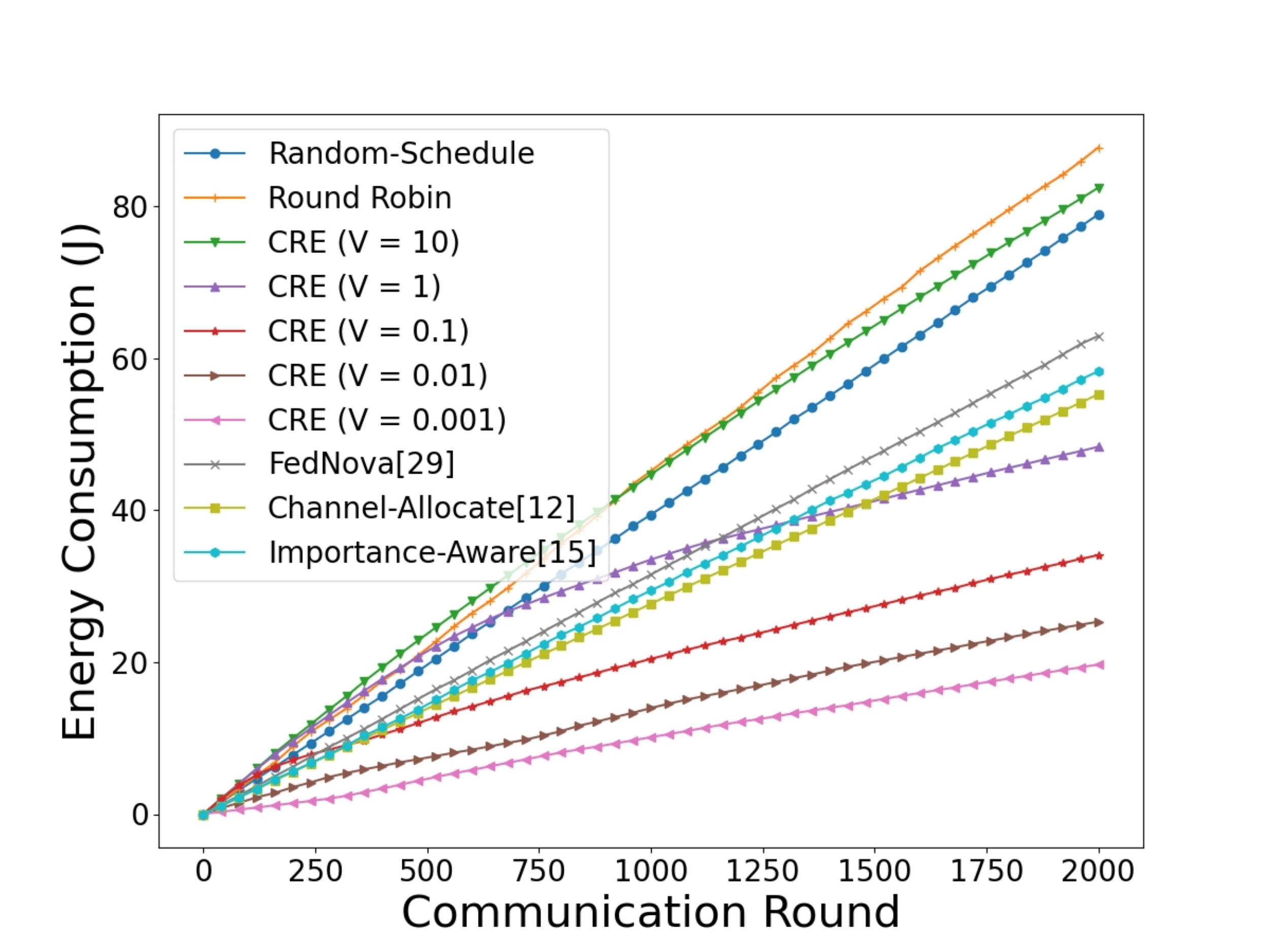}}
  \hspace{0in}
  \caption{\textcolor{black}{Test accuracy and accumulated energy consumption curves of the proposed algorithm using various $V$ with $d = 0.4$ and $\sigma = 100$.}}
  \label{figure: Energy}
\end{figure*}
For the neural network setting, the multilayer perceptron (MLP) consists 50 neurons in the hidden layer and the cross entropy loss function is applied to accomplish the handwritten digit identification, i.e. the MNIST dataset \cite{MNIST}. For another colored image identification, i.e. the CIFAR-10 dataset \cite{cifar10}, we employ a convolutional neuron network (CNN) consisting of two convolution layers with 64 $5\times5$ kernels and three hidden layers with 1024, 384, 192 neurons, repectively. The experimentation results show that our CRE optimization framework also works well for models (e.g., the neural network) whose loss functions are non-convex.
\par
The data setting is described as follows. In general, the dataset size $D_i$ of each client follows the Gaussian distribution $N(\mu, \sigma^2)$. In our simulations, $\mu$ is set as 1000 and $\sigma$ varies. Besides, non-IID datasets are generated by the method in \cite{non_iid_data}. \textcolor{black}{To be specific, the whole training set is divided into two parts: the common set contains common data and the particular set contains particular data. The particular set is further divided into particular subsets based on the requirement.} And the dataset of client $i$ is also divided into the particular dataset $\mathcal D^{\rm par}_i$ and the common dataset $\mathcal D^{\rm com}_i$. One particular dataset of each client corresponds to one particular subset. Take 10 clients as an instance, the particular set can be divided into 10 subsets, and each subset consists samples of one class. Thus, every sample of $\mathcal D^{\rm par}_i$ is sampled randomly from the corresponding particular subset. And $\mathcal D^{\rm com}_i$ is a disjoint subset of the common set. Finally, the whole dataset is $\mathcal D_i = \mathcal D_i^{\rm par} \bigcup D_i^{\rm com}$. The non-IID degree is defined by $d_i \triangleq \frac{|\mathcal D_i^{\rm par}|}{D_i}$. Hence, the global non-IID degree is obtained by taking the weighted average of dataset sizes, that is $d = \sum^N_i w_i d_i$. \par
For comparison purpose, two basic baselines in \cite{scheduling_baseline} are presented: (a) random scheduling algorithm that allocates channels to clients randomly, (b) round robin algorithm that allocates channels to clients in rotation. Our paper proposes: (c) CRE algorithm that solves our optimization problem. To demonstrate the benefit of CREs, other 3 algorithms are included: \textcolor{black}{(d) FedNova algorithm in \cite{different_epoch} that normalizes local models with different local epochs.} (e) channel-allocate algorithm in \cite{importance_of_local_updates} that only optimizes client scheduling and channel allocation with a fixed number of local epochs. \textcolor{black}{(f) importance-aware algorithm in \cite{importance-aware} optimize client scheduling with fixed channel allocation and a fixed number of local epochs.}\par
\begin{table*}[t]
  \centering
  \caption{\textcolor{black}{Accuracy on Datasets}}
  \begin{tabular}{c|c|c|cccccc}
  \hline \hline
  Model & \multicolumn{2}{c|}{Heterogeneity} & \multicolumn{6}{c}{Curve Type} \\ \cline{2-9}
  and & \multirow{2}{*}{$d$} & \multirow{2}{*}{$\sigma$} & Random & Round & \multirow{2}{*}{CRE} & \multirow{2}{*}{FedNova} & Channel & Importance \\
  Dataset & & & -Schedule & Robin & & & -Allocate & -Aware \\ \hline
  & \multirow{3}{*}{0.2} & 50 & 82.66\%                          & 82.50\% & \textbf{86.29\%} & 84.32\% & 85.09\% & 84.81\% \\
  & & 100 & 82.59\% & 82.76\% & \textbf{86.58\%} & 84.25\% & 84.96\% & 84.93\% \\
  &  & 150   & 80.85\%  & 81.09\% & \textbf{86.13\%} & 83.24\% & 85.06\% & 83.89\%  \\
  MLP & \multirow{3}{*}{0.4} & 50 & 81.76\%  & 81.97\%  & \textbf{86.15\%} & 83.79\% & 85.42\%  & 83.07\% \\
  on &   & 100 & 81.95\% & 81.50\%  & \textbf{85.83\%}  & 83.92\% & 85.01\% & 82.75\%  \\
  MNIST &  & 150  & 81.50\%  & 81.37\% & \textbf{85.64\%} & 83.26\%  & 84.47\% & 81.98\%  \\
  & \multirow{3}{*}{0.6} & 50  & 81.76\%  & 81.89\%  & \textbf{85.80\%} & 83.31\%  & 84.84\%  & 82.03\% \\
  &   & 100   & 82.13\%   & 81.97\%  & \textbf{85.30\%}  & 83.34\%  & 81.27\%                           & 81.17\% \\
  & & 150 & 81.36\%   & 81.33\%   & \textbf{85.03\%}  & 83.03\%  & 78.92\% & 77.82\% \\ \hline
  & \multirow{3}{*}{0.2} & 50 & 53.20\%  & 53.58\%  & \textbf{58.97\%}     & 57.26\%   & 57.56\% & 56.80\%   \\
  &   & 100  & 52.95\%   & 53.55\% & \textbf{58.86\%}     & 57.09\%  & 57.62\%   & 56.09\% \\
  &   & 150  & 53.02\% & 53.46\%   & \textbf{59.06\%}     & 57.30\%                  & 57.47\%   & 55.41\%   \\
  CNN & \multirow{3}{*}{0.4} & 50   & 50.60\%   & 51.12\%   & \textbf{56.47\%}     & 54.59\%  & 53.92\%   & 53.42\%  \\
  on &   & 100 & 51.99\%  & 52.50\%  & \textbf{56.41\%}  & 54.72\%  & 54.00\%                           & 52.70\%  \\
  CIFAR-10&   & 150   & 52.02\%  & 52.35\%  & \textbf{56.38\%} & 54.81\% & 51.91\% & 50.29\%  \\
  & \multirow{3}{*}{0.6} & 50 & 47.12\%  & 48.22\% & \textbf{53.14\%}  & 51.84\% & 50.46\% & 49.60\%  \\
  &  & 100 & 47.31\%  & 48.20\% & \textbf{52.43\%} & 51.35\% & 48.28\%                           & 47.48\% \\
  &   & 150  & 46.24\%   & 47.86\%   & \textbf{52.30\%}   & 50.61\%  & 46.22\%     & 43.77\% \\ \hline \hline
  \end{tabular}
  \label{table: Datasets}
\end{table*}
In our objective function in \textbf{P2}, both the performance of the model and the energy consumption are considered and a penalty factor $V$ takes a trade-off between them. To choose a proper factor, different values of $V$, i.e., $V = 10, 1, 0.1, 0.01, 0.001$ are used in \ref{subsection: Energy}. After finding the proper factor, detailed results of the performance for different heterogeneities are presented in \ref{subsection: MNIST} and \ref{subsection: CIFAR10}. In our simulations, all curves are obtained by taking the average of 5 experiment results. \par
\subsection{Trade-off between Performance and Energy Consumption \label{subsection: Energy}}
Fig. \ref{figure: Energy} shows how $V$ affects the testing accuracy and the energy consumption. From Fig. \ref{figure: Energy}(a) and (b), it can be observed that as $V$ increases, accuracies of CREs become lower and energy cosumptions decrease, respectively. This is because large $V$ emphasizes the learning performance rather than energy consumption, and vice versa. In particular, the algorithm with $V = 10$ achieves the highest accuracy but consumes even more energy than the round robin algorithm. On the other hand, although the algorithm with $V = 0.001$ consumes the least energy, its accuracy is much worse than other 5 baselines. This reveals that choosing proper $V$ is a key factor to trade off the learning performance and the energy consumption. In Fig. \ref{figure: Energy}(a), another observation is that the accuracy gain gets smaller as $V$ increases. It attributes to the fact that limited wireless resources restrict the performance of FL. \textcolor{black}{In Fig. \ref{figure: Energy}(b), we can see that the FedNova algorithm consumes more energy than the channel-allocate algorithm and the importance-aware algorithm. It is due to the fact that there is no design for wireless communication, which leads to a waste of energy consumption. Since the energy consumption of CRE $(V=0.1)$ is just less than other baselines, we choose $V = 0.1$ to accomplish following simulations on the MNIST dataset.} Despite only a result with $d = 0.4$ and $\sigma = 100$, the conclusion is universal since the data heterogeneity can not affect the overall energy consumption. From Fig. \ref{figure: Energy}(b), it is also observed that energy consumption slope decreases as communication rounds increases and a turning point appears. Actually, the benefit of training models decreases as FL converges. In this case, our CRE algorithm turns to save energy to pursue a larger benefit, thus the curve slope decreases. \par
\textcolor{black}{Similarly, Fig. \ref{figure: Energy}(c) and (d) validate above conclusions about $V$ with CNN on the CIFAR-10 dataset. Hence, $V = 1$ is chosen for later simulations on the CIFAR-10 dataset.}

\begin{figure}[t]
  \centering
  \subfigure[$d = 0.2$, $\sigma = 50$ ]{\includegraphics[width=0.24\textwidth]{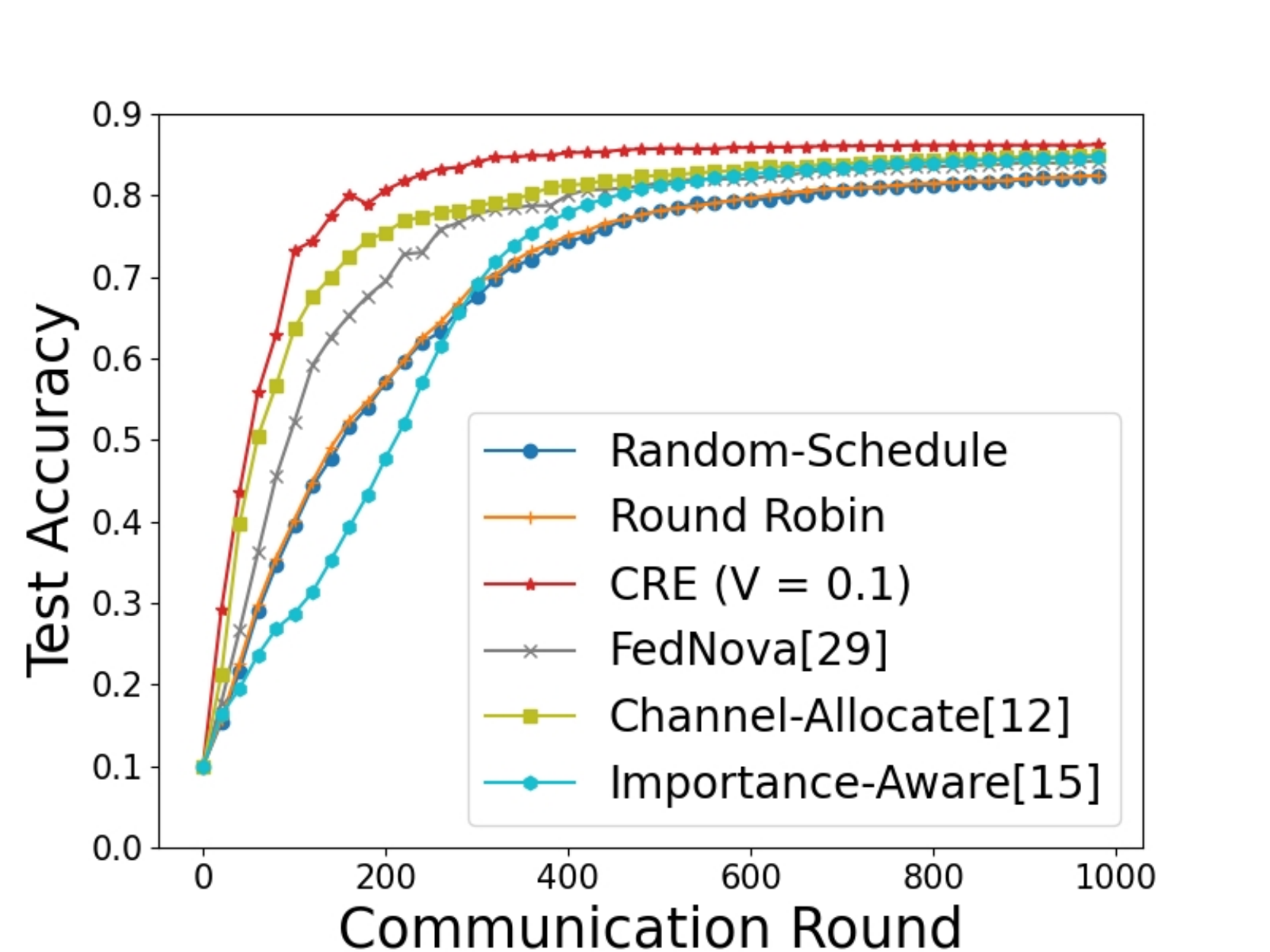}}
  \subfigure[$d = 0.2$, $\sigma = 150$ ]{\includegraphics[width=0.24\textwidth]{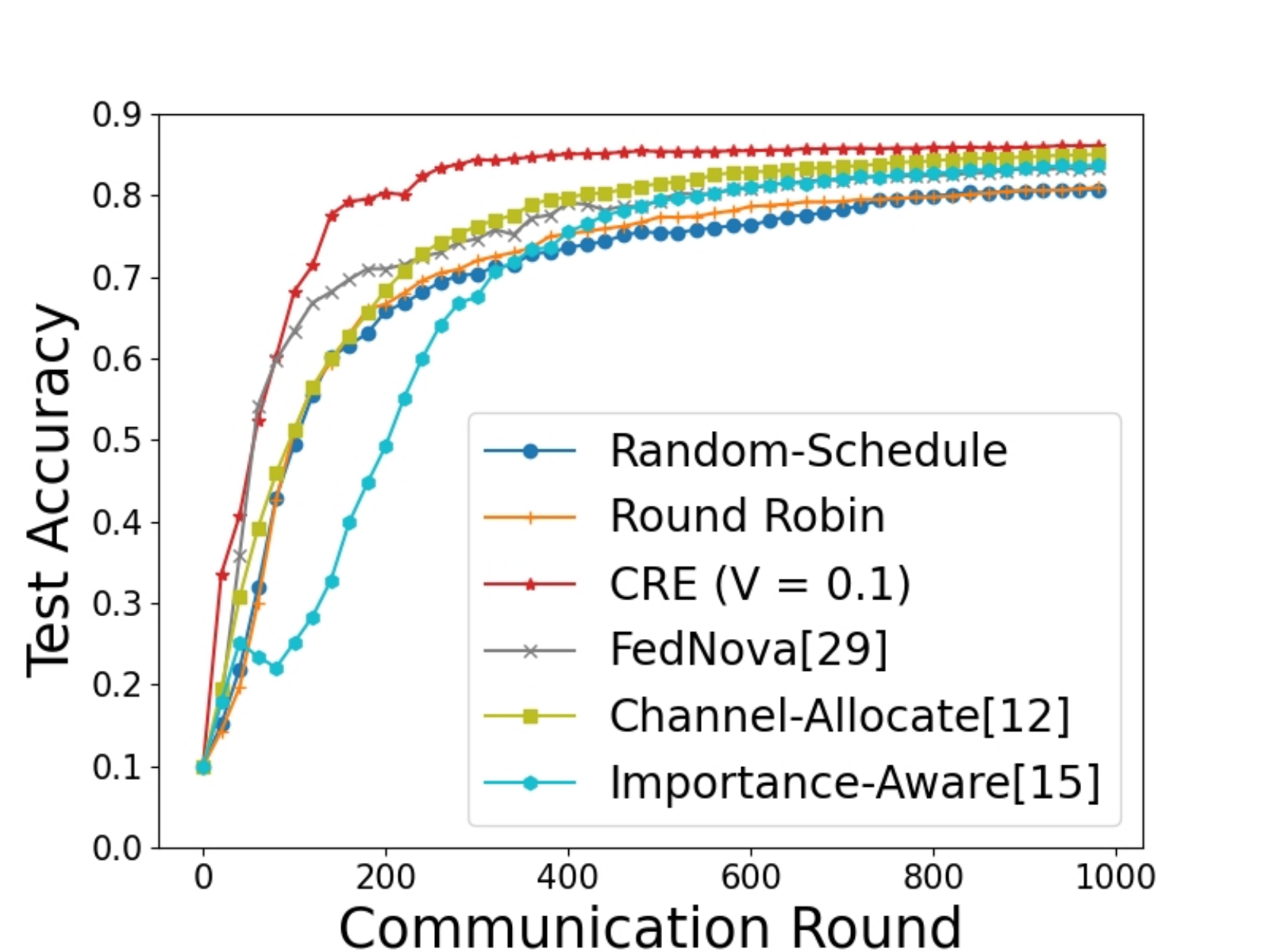}}
  \\
  \subfigure[$d = 0.6$, $\sigma = 50$ ]{\includegraphics[width=0.24\textwidth]{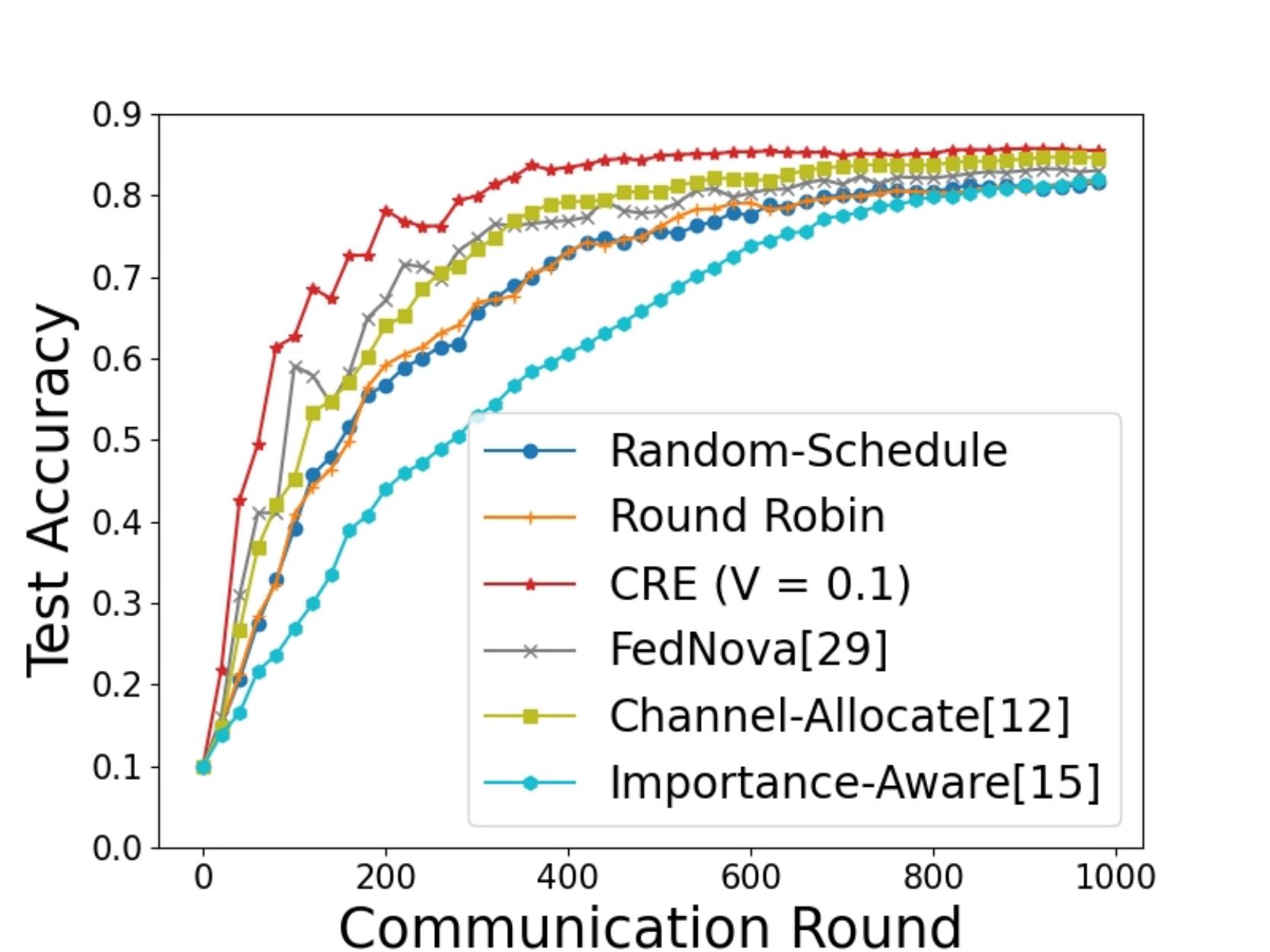}}
  \subfigure[$d = 0.6$, $\sigma = 150$ ]{\includegraphics[width=0.24\textwidth]{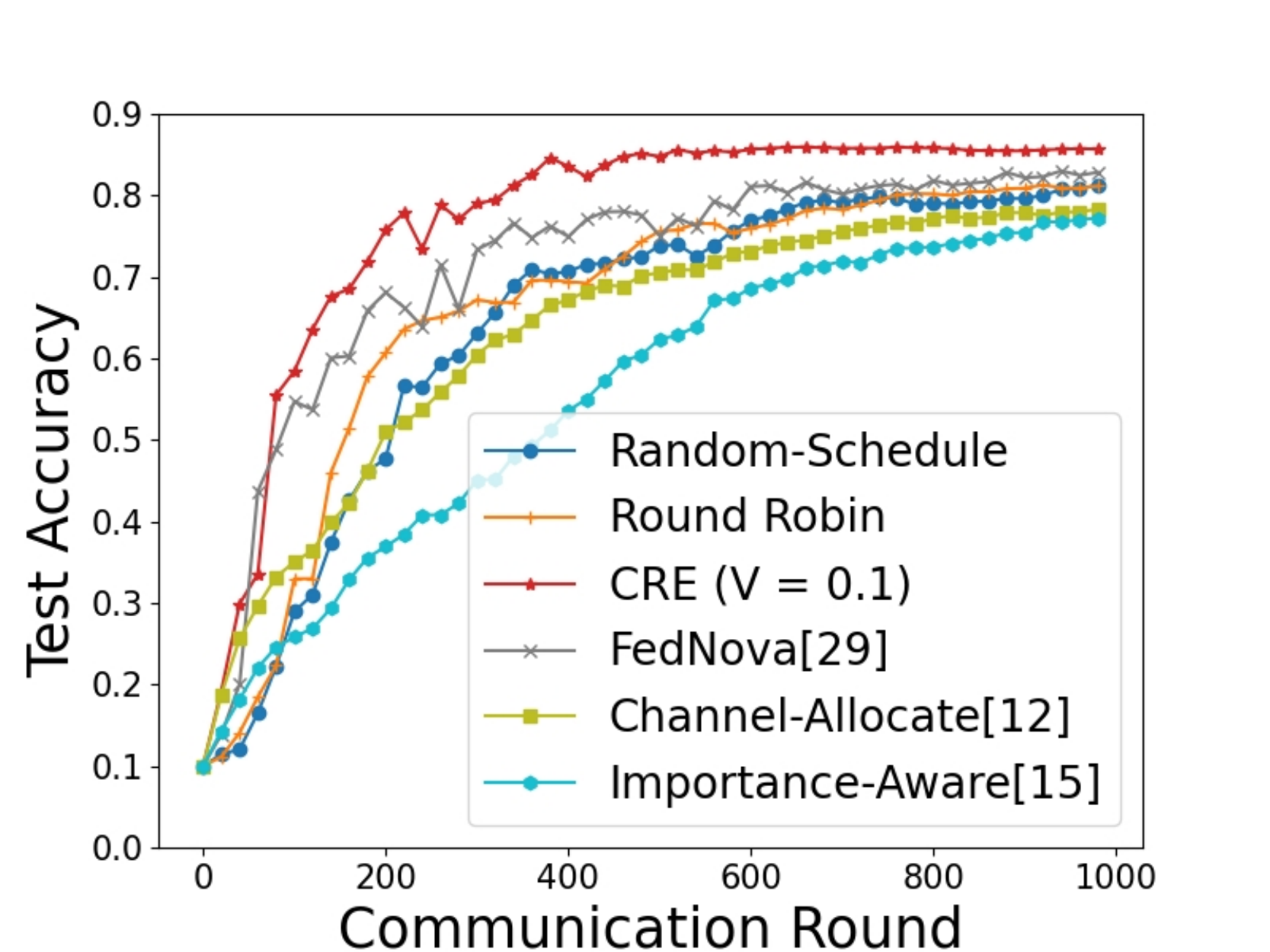}}
  \caption{\textcolor{black}{Test accuracy curves of different 6 algorithms with $d = 0.2, 0.6$ and $\sigma = 50, 150$ on the MNIST dataset.}}
  \label{figure: MNIST}
\end{figure}
\subsection{Handwritten Digit Identification\label{subsection: MNIST}}
Fig. \ref{figure: MNIST} depicts the learning performance of our CRE algorithm as compared with 5 baselines on the MNIST dataset with different heterogeneities. It can be observed from all subfigures that our CRE algorithm achieves the highest accuracy and the fastest convergence. \textcolor{black}{The underlaying reason is that our CRE algorithm optimizes FL performance with a joint consideration of communication
constraints and the data heterogeneity. From Fig. \ref{figure: MNIST}(a), (b), it can be seen that the channel-allocate algorithm and the importance-aware algorithm perform better than basic baselines and the FedNova algorithm. It is due to the fact that these two algorithms optimize the performance of wireless FL and can handle the scene with low heterogeneity. However in Fig. \ref{figure: MNIST}(d), it is obvious that the two algorithms both get worse, while our proposed CRE algorithm is still the best. The reason is that the two algorithms can not tackle the high heterogeneity without acquiring the characteristic of data heterogeneity. The FedNova algorithm, by contrast, keeps stable performance in Fig. \ref{figure: MNIST}(a)-(d). However, due to lack of optimization of communication, the performance of the FedNova algorithm can not reach the performance of our CRE algorithm.} \par
\begin{figure}[t]
  \centering
  \subfigure[$d = 0.2$, $\sigma = 50$ ]{\includegraphics[width=0.24\textwidth]{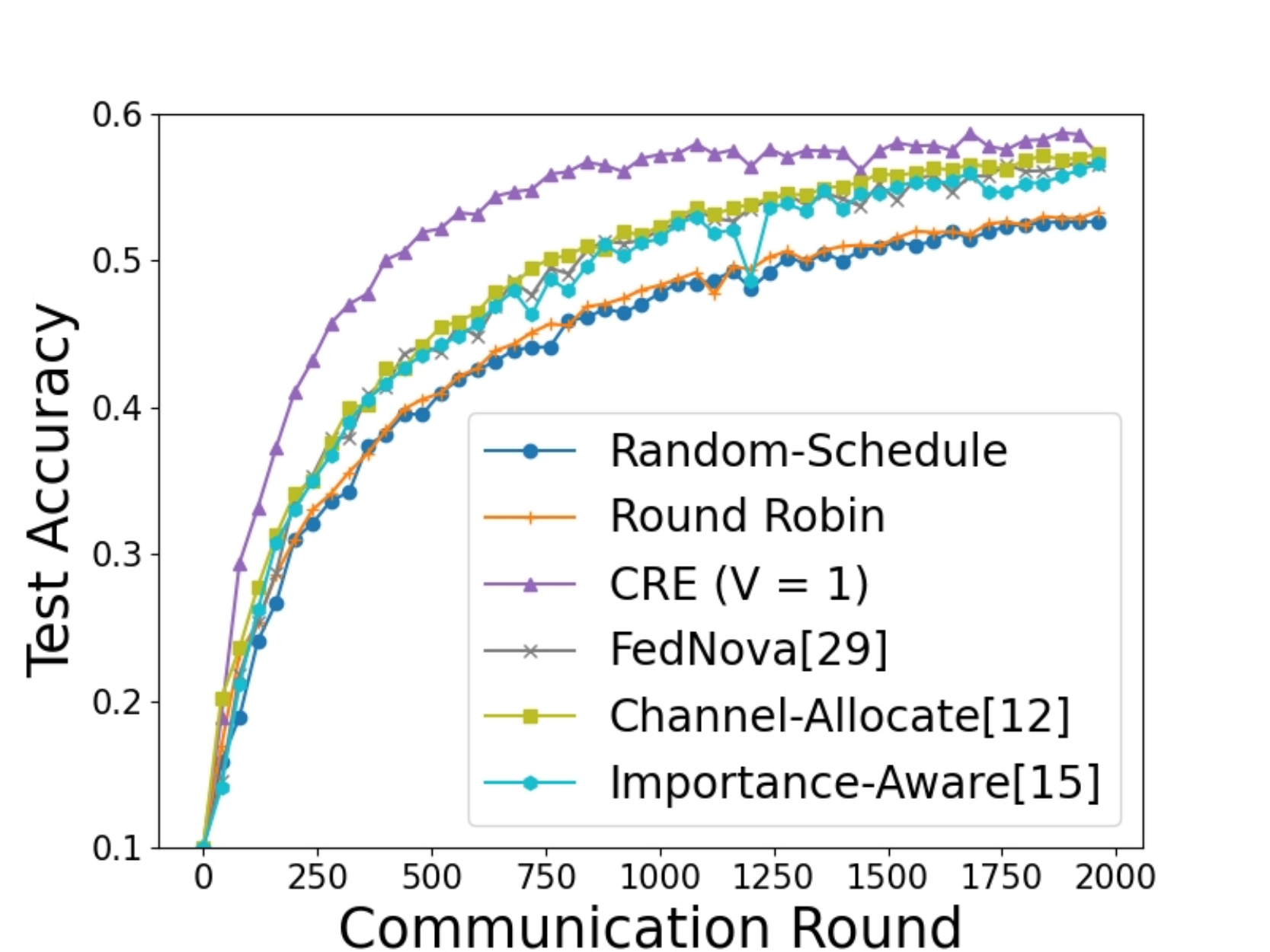}}
  \subfigure[$d = 0.2$, $\sigma = 150$ ]{\includegraphics[width=0.24\textwidth]{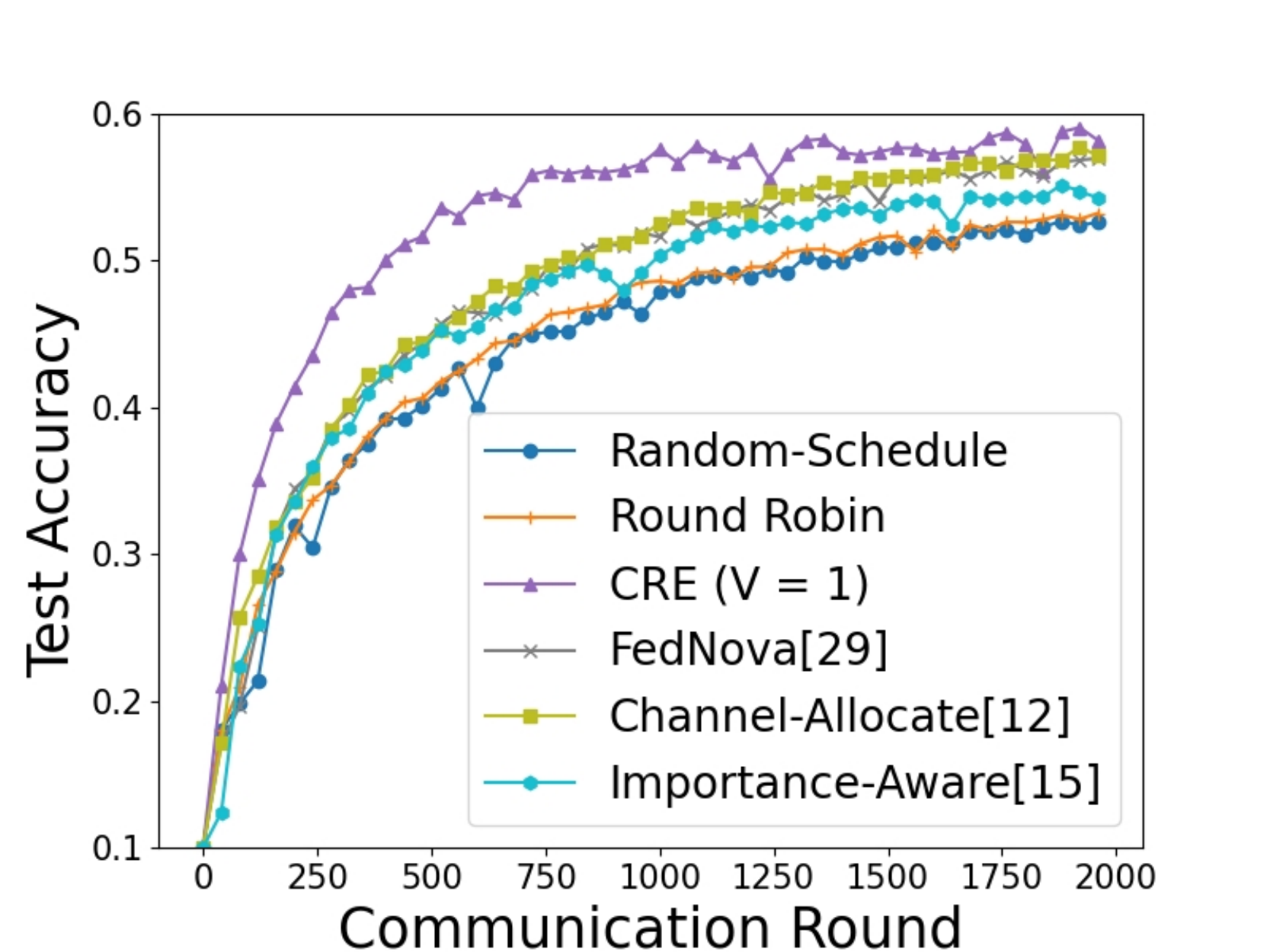}}
  \\
  \subfigure[$d = 0.6$, $\sigma = 50$ ]{\includegraphics[width=0.24\textwidth]{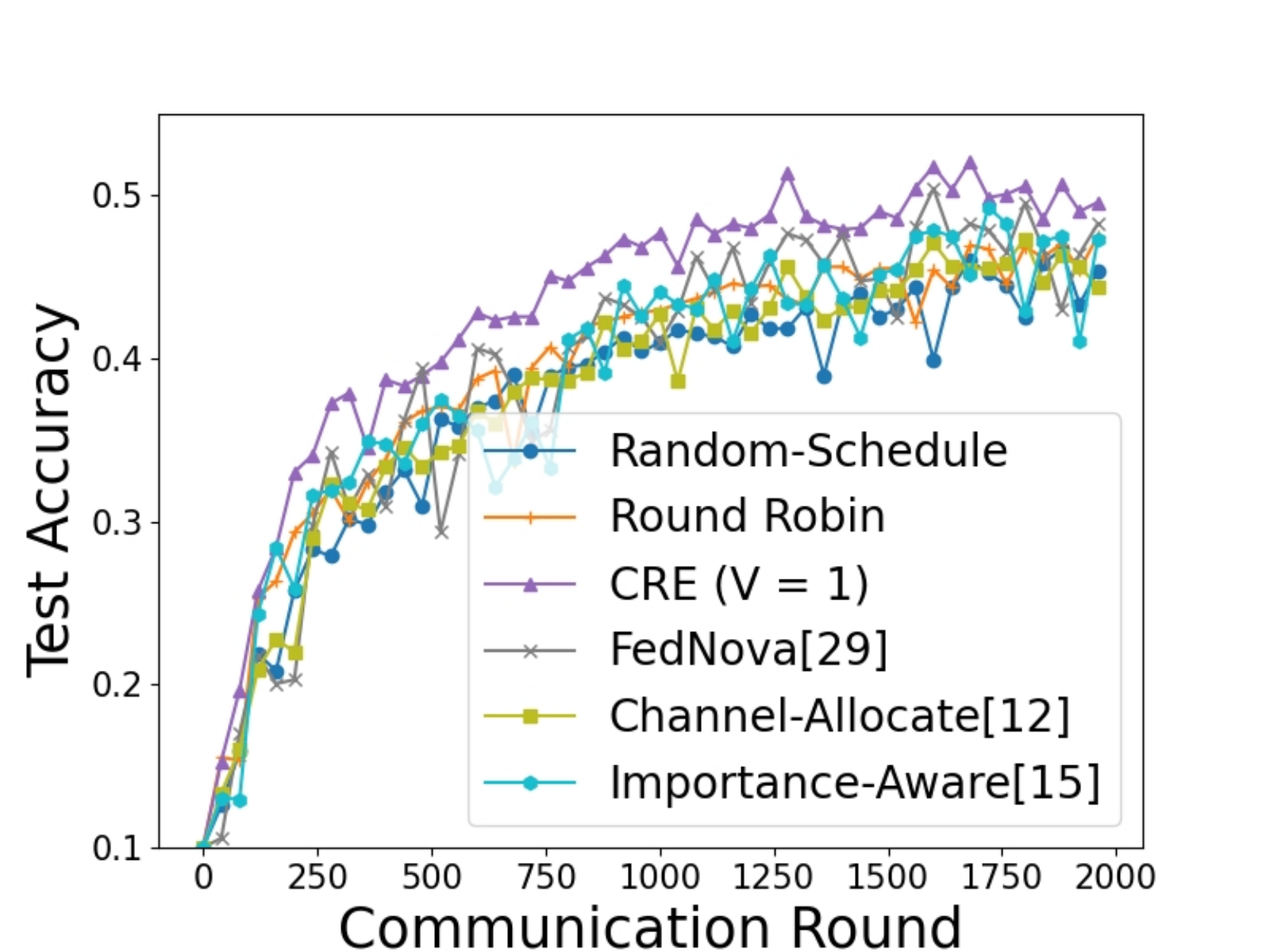}}
  \subfigure[$d = 0.6$, $\sigma = 150$ ]{\includegraphics[width=0.24\textwidth]{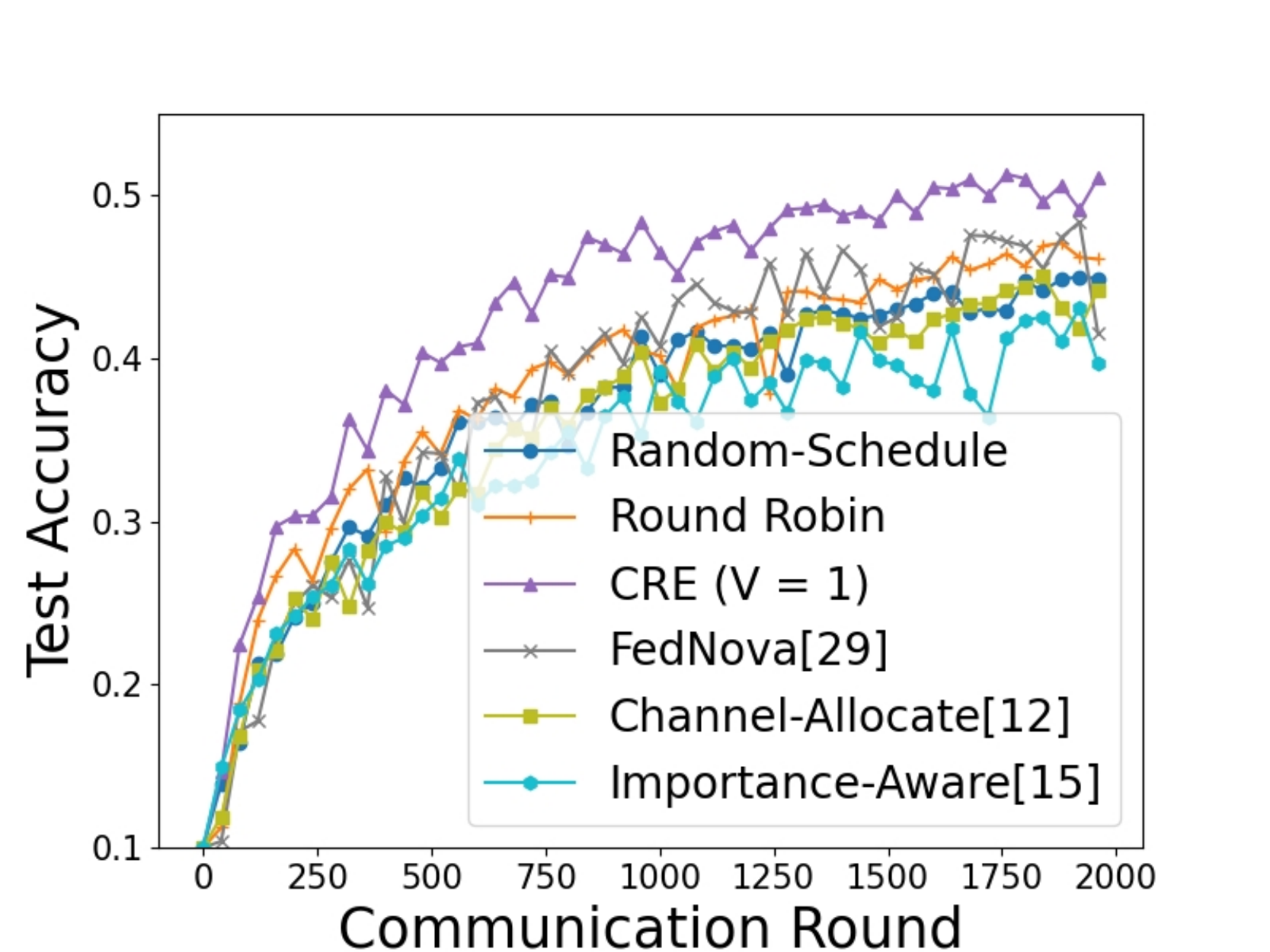}}
  \hspace{0in}
  \caption{\textcolor{black}{Test accuracy curves of different 6 algorithms with $d = 0.2, 0.6$ and $\sigma = 50, 150$ on the CIFAR-10 dataset.}}
  \label{figure: CIFAR10}
\end{figure}
Fig. \ref{figure: MNIST} also shows that as the non-IID degree increases, all curves get fluctuant and converge slowly. These phenomena are universal for FL on non-IID data. Our CRE algorithm still performs best as shown in the top half of Table \ref{table: Datasets}.
\begin{table*}[ht]
\centering
\caption{\textcolor{black}{Detailed Solutions of Our CRE Algorithm over A Period of Time}}
\begin{tabular}{c|cc|cc|cc|c}
\hline \hline
Communication & \multicolumn{2}{|c|}{Channel 1} & \multicolumn{2}{|c|}{Channel 2} & \multicolumn{2}{|c|}{Channel 3} & Local \\
\cline{2-7}
Round & Client   & Uplink Power  & Client   & Uplink Power  & Client   & Uplink Power  & Epoch \\ \hline
201 & 7 & 27.24 mW   & 4  & 199.53 mW & 0   & 0    & 5      \\
202 & 10  & 2.64 mW  & 0        & 0  & 2 & 199.53 mW & 5     \\
203 & 4 & 44.66 mW & 10 & 199.53 mW & 9 & 10.62 mW & 4 \\
204 & 3 & 199.53 mW & 4 & 120.59 mW & 0 & 0 & 4 \\
205  & 0  & 0   & 6 & 199.53 mW   & 8  & 82.02 mW  & 4  \\
206   & 7 & 199.53 mW  & 1  & 78.73 mW  & 4  & 77.71 mW  & 3 \\
207 & 4 & 199.53 mW  & 0   & 0   & 8  & 108.17 mW   & 5  \\
208   & 2  & 199.53 mW & 9   & 39.37 mW   & 0  & 0   & 5  \\
209  & 4  & 199.53 mW  & 6  & 152.39 mW & 0  & 0  & 6  \\
210   & 0  & 0   & 4  & 66.56 mW   & 6   & 199.53 mW   & 4 \\ \hline \hline
\end{tabular}
\label{table: solution}
\end{table*}

\begin{table*}[ht]
  \centering
  \caption{\textcolor{black}{Statistical Data of Scheduling Numbers and Local Epoch Numbers}}
\begin{tabular}{c|c|cccccccccc|c}
\hline \hline
\multirow{2}{*}{Curve   Type} & Client & 1    & 2    & 3    & 4    & 5    & 6    & 7    & 8    & 9    & 10   & \multirow{2}{*}{SUM} \\ \cline{2-12}
& DataSet Size      & 1091 & 1017 & 1257 & 1039 & 1004 & 876  & 920  & 1025 & 1069 & 1054 &                      \\ \hline
\multirow{2}{*}{CRE (V=1)}                  & Schedule & 380 & 413  & 345  & 375  & 422  & 461  & 446  & 400  & 363  & 443  & 4048 \\
& Epoch & 1179 & 1251 & 1037 & 1209 & 1378 & 1503 & 1423 & 1212 & 1109 & 1419 & 12720                \\  \hline
Importance & Schedule & 29   & 375  & 42   & 526  & 690  & 1294 & 1097 & 197  & 124  & 292  & 4666                 \\
-Aware \cite{importance-aware} & Epoch & 84   & 1122 & 133  & 1575 & 2067 & 3879 & 3288 & 588  & 369  & 873  & 13978                \\ \hline
Random & Schedule & 581  & 528  & 539  & 566  & 565  & 561  & 517  & 580  & 533  & 563  & 5533  \\
Scheduling & Epoch & 1219 & 1077 & 1072 & 1126 & 1164 & 1190 & 1053 & 1216 & 999  & 1117 & 11233 \\ \hline \hline
\end{tabular}
\label{table: statistic}
\end{table*}
\subsection{Colored Image Identification\label{subsection: CIFAR10}}
Fig. \ref{figure: CIFAR10} shows that the learning performance of all algorithms on the CIFAR-10 dataset with different heterogeneities. From Fig. \ref{figure: CIFAR10}, we can observe that our CRE algorithm achieves the highest accuracy and the fastest convergence, \textcolor{black}{In Fig. \ref{figure: CIFAR10}(d), it can be seen that the channel-allocate algorithm and the importance-aware algorithm are worst among all algorithms.} These conclusions are similar to those with MLP on the MNIST dataset, which suggests our algorithm can be generalized to other neural networks and datasets. In Fig. \ref{figure: CIFAR10}(c) and (d), it is shown that a large fluctuation occurs due to non-IID data. In addition, the bottom half of the Table \ref{table: Datasets} shows that all algorithms get noticeably worse, when $d=0.6$ and $\sigma = 150$. This is caused by characteristics of the CIFAR-10 dataset. Nevertheless, our CRE algorithm achieves more than 5\% gain compared with common wireless optimization algorithms.
\begin{figure}[t]
  \centering
  \subfigure[Loss of MLP on MNIST ]{\includegraphics[width=0.24\textwidth]{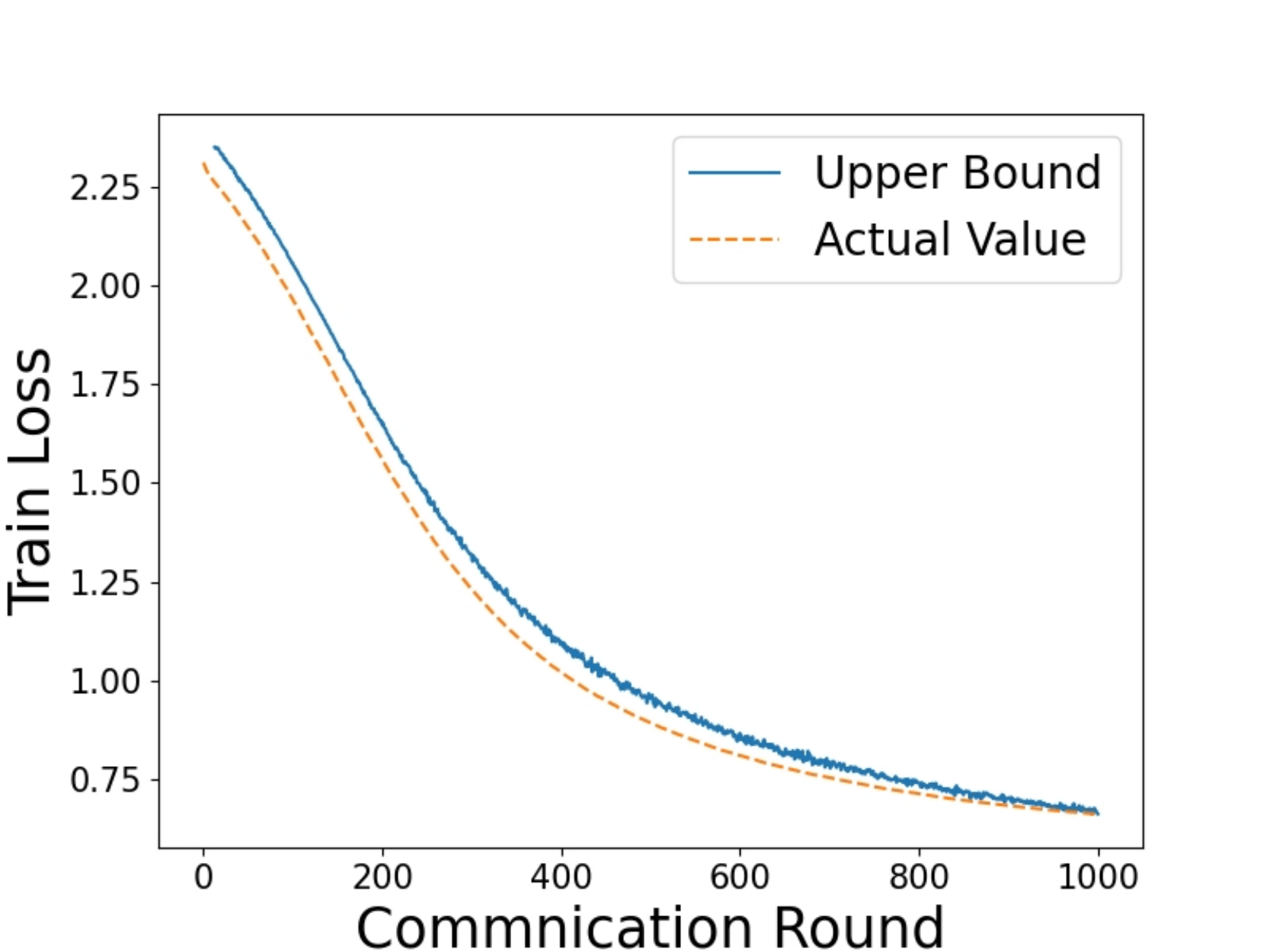}}
  \subfigure[Loss of CNN on CIFAR-10 ]{\includegraphics[width=0.24\textwidth]{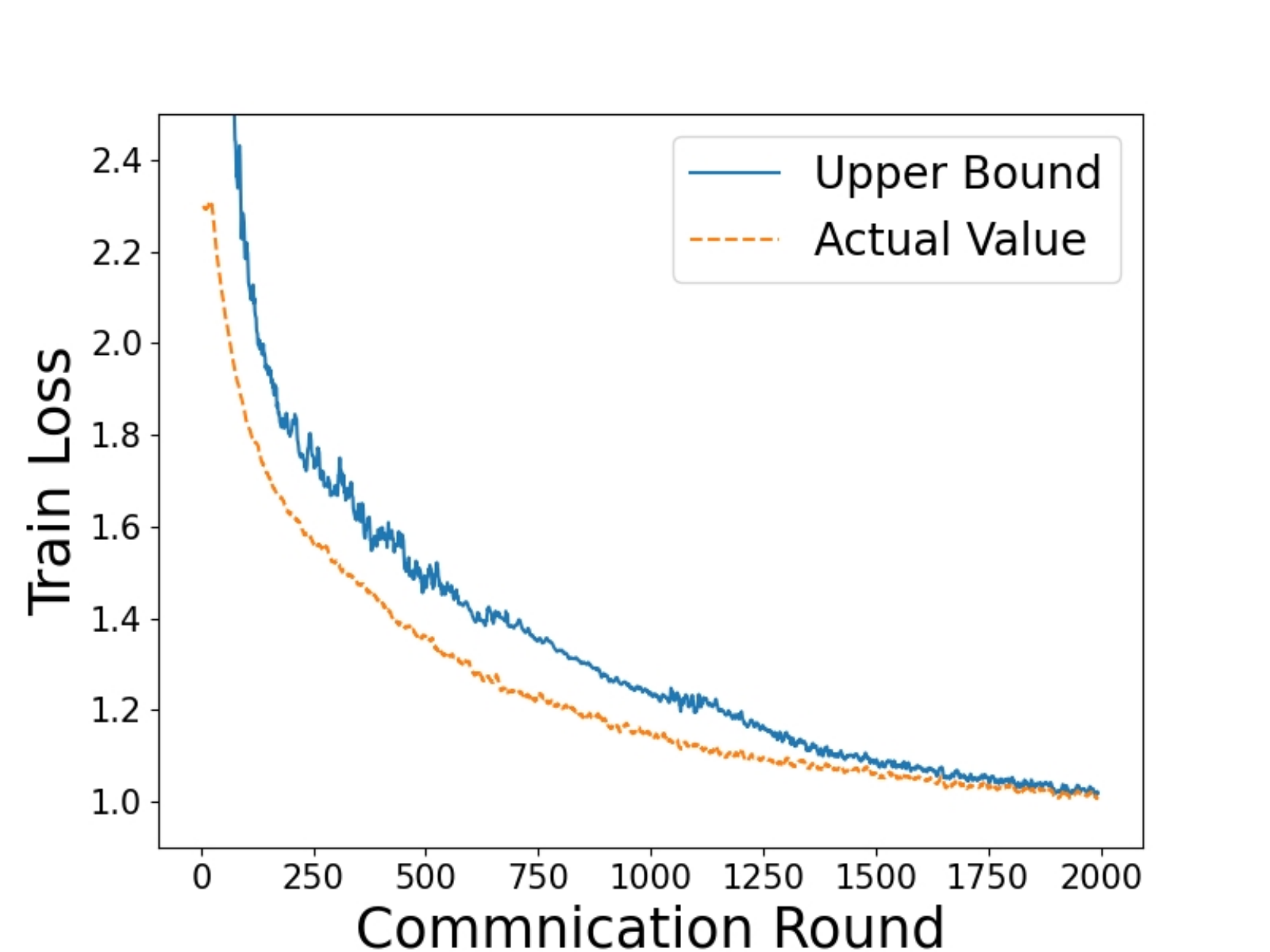}}
  \caption{\textcolor{black}{Derived upper bounds and the actual loss function on training datasets.}}
  \label{figure: Bound}
\end{figure}
\textcolor{black}{
\subsection{Perfomrnace Verification\label{subsection: Verification}}
Fig. \ref{figure: Bound} shows the comparison between the derived upper bounds and the actual loss functions. In our CRE algorithm, the derived upper bound, as a substitute of the loss function which can not be predicted before training, is the objective function. Thus, Fig. \ref{figure: Bound} validates that the derived upper bound can characterize the trend of the loss function with a small gap and the gap vanishes to 0 as the training process. This is because model property parameters of upper bounds are not accurately estimated in the initial process. And in the later process, the model convergence refines the estimation of model property parameters and the upper bound becomes precise.}\par
\textcolor{black}{
Table \ref{table: solution} and \ref{table: statistic} provides detailed solutions and statistical data of one experiment with $d = 0.4, \sigma = 100$ on the CIFAR-10 dataset. In Table \ref{table: solution}, it is observed that client 1-10 are all scheduled and 3 available channels are allocated within 201-210 communication rounds. This result demonstrates fairness among clients and channels. From another perspective, all 3 channels are not always used in a communication round. The underlaying reason is that not all channel states are suitable for fast uplink communication after a large number of local epochs, such as 6 in the 209 communication round. Table \ref{table: statistic} illustrates that despite a fewer numbers of participating, our CRE solution has more epochs compared to the random scheduling algorithm. Moreover, conventional optimization methods, such as the importance-aware algorithm in \cite{importance-aware}, pursue more training epochs at the cost of fairness among clients. Such a tendency seriously degrades the performance when $d$ is high. To sum up, our CRE algorithm keeps basic fairness of scheduling for each client, increases scheduling clients with small datasets suitably, and adjusts the number of local epochs adaptively. This is why our CRE algorithm performs best on heterogeneous data.} \par
\begin{figure}[ht]
  \centering
  \subfigure[Accuracy on MNIST ]{\includegraphics[width=0.24\textwidth]{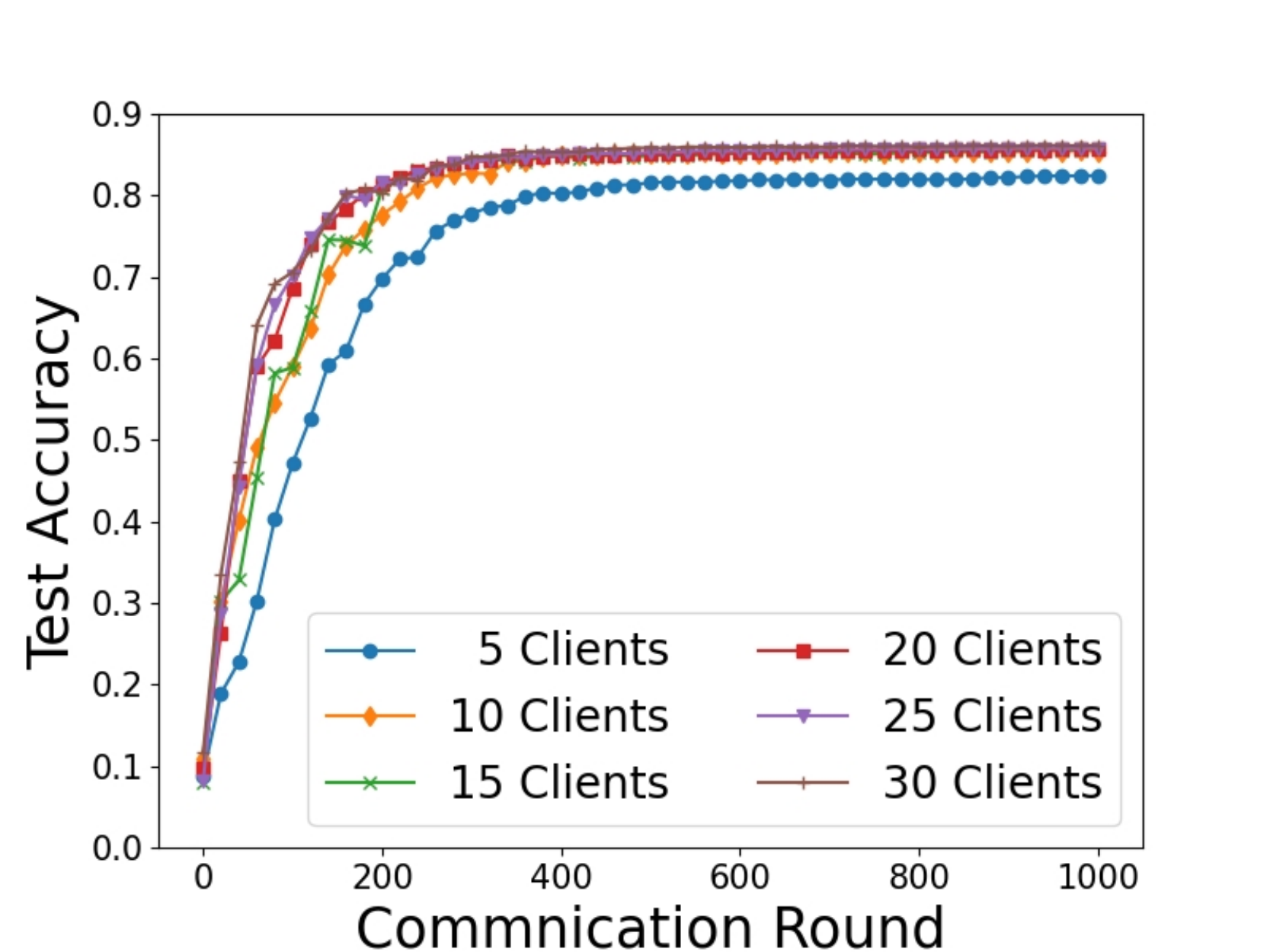}}
  \subfigure[Energy on MNIST ]{\includegraphics[width=0.24\textwidth]{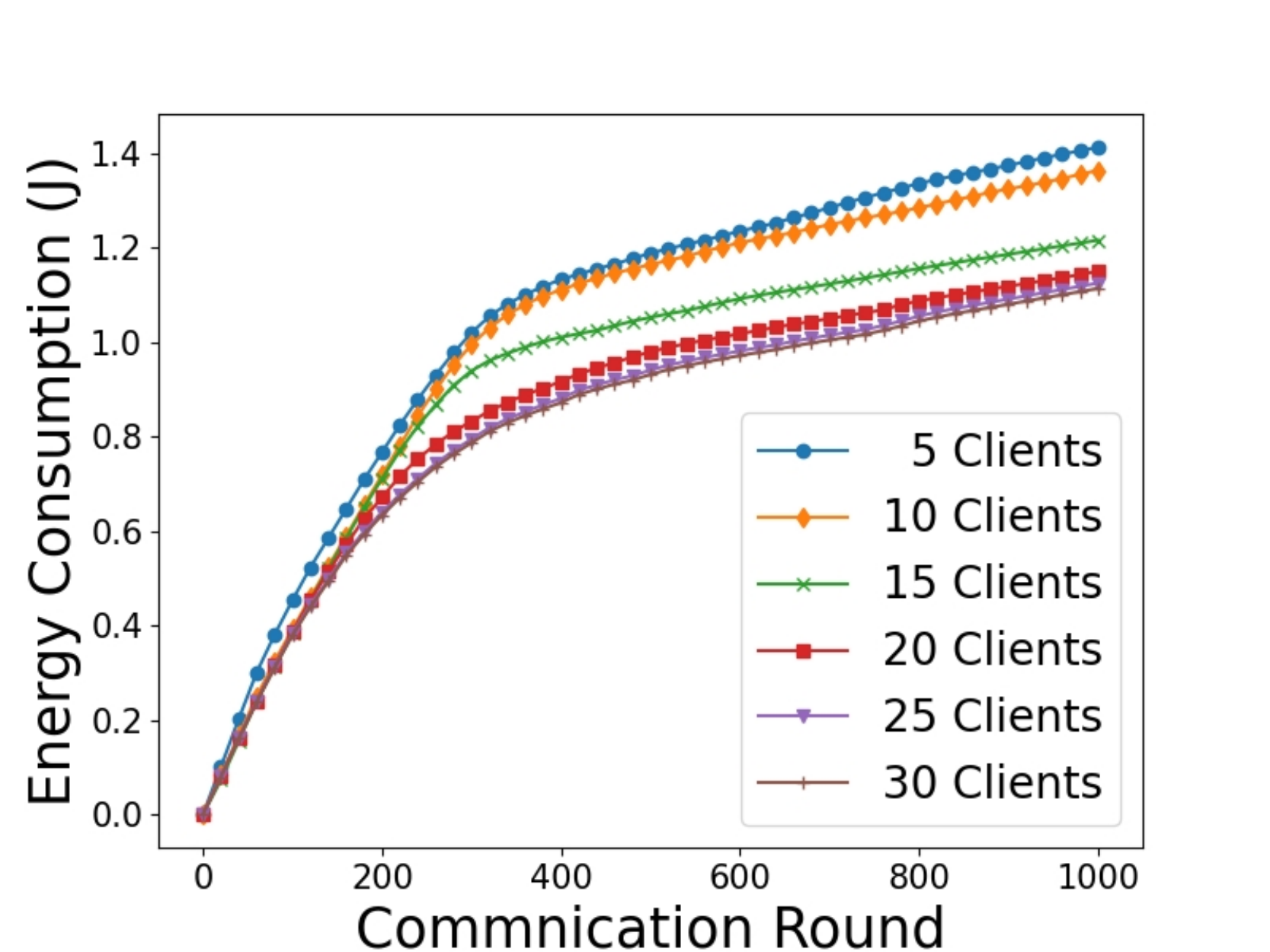}} \\
  \subfigure[Accuracy on CIFAR-10 ]{\includegraphics[width=0.24\textwidth]{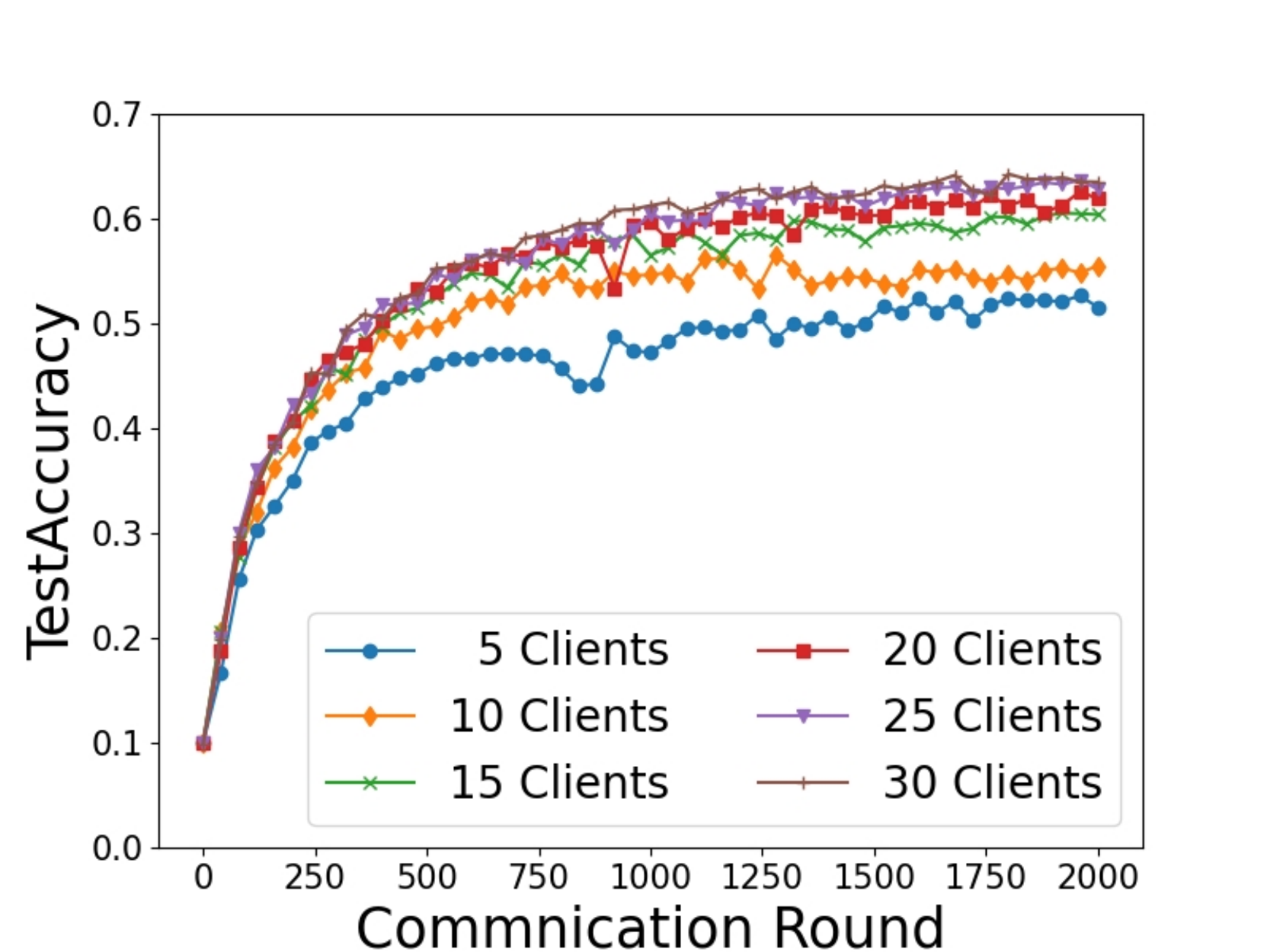}}
  \subfigure[Energy on CIFAR-10 ]{\includegraphics[width=0.24\textwidth]{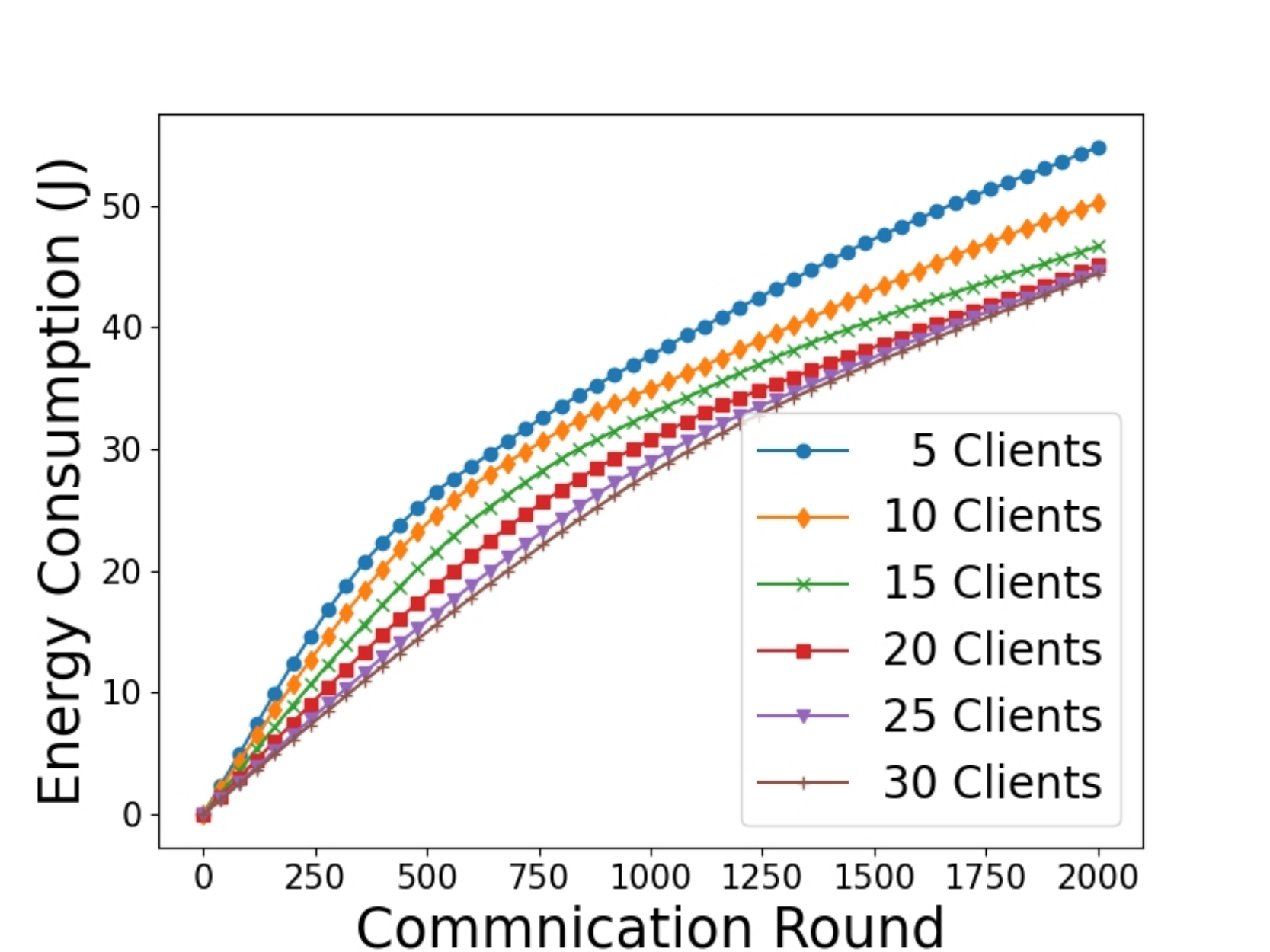}}
  \hspace{0in}
  \caption{\textcolor{black}{Test accuracy and energy consumption curves of CRE (V=0.1) algorithm for different clients.}}
  \label{figure: Client}
\end{figure}
\textcolor{black}{
Fig. \ref{figure: Client} depicts how the number of clients influences FL performance and energy consumption. It is noticed that as the number of clients increases, FL performance gets better and the energy consumption gradually decreases. These improvements attribute to more datasets and more channel states to choose, which are consequences of more clients. With the larger feasible region, our CRE algorithm can further pursue high performance and low energy consumption. Such a conclusion validates our CRE algorithm in the scene with more clients. When the number of clients reaches 30, improvements of more clients are few. This is due to the fact that constraints on FL performance and energy consumption are primarily influenced by the scarcity of channels rather than the increasing number of clients.
}
\section{Conclusion}
In this paper, we have proposed an optimization problem of CREs for FL with different data sizes and non-IID data over wireless networks. First, an optimization problem of client scheduling, channel allocation, uplink power control, and number of local epochs design has been formulated to minimize the final loss function. To characterize the loss function, we have developed a closed-form upper bound with the dataset size vector and the data divergence vector. Moreover, we have proposed a method to estimate model property parameters and the data divergence vector. Then, by means of Lyapunov technique, the formulated problem for the whole process has been transformed into a one-communication-round problem, which can be solved by using Tammer decomposition. Our simulation results have demonstrated that the proposed algorithm works well for solving data heterogeneity among the clients. Compared to baselines, our method \textcolor{black}{has} achieved the best accuracy with lower energy consumption. In the future, other \textcolor{black}{heterogeneous} characteristics of clients such as CPU efficiency and battery power will be taken into consideration. Also, different numbers of local epochs for individual clients will be explored.
\appendix
Since the derivation for all communication rounds are same, the superscript $n$, which represents the $n$-th communications round, is omitted throughout the appendix.
\subsection{Proof of Theorem \ref{theorem: parameter difference}\label{subsection: proof of theorem 1}}
To prove \textbf{Theorem \ref{theorem: parameter difference}}, \textbf{Lemma \ref{lemma: local parameter difference}} is proposed as follows.
\begin{lemma}
The difference between $\boldsymbol\theta_i^{m}$ and $\boldsymbol\phi^{m}$ is bounded by
\begin{equation}
\begin{aligned}
& \| \boldsymbol\theta_i^{m} - \boldsymbol\phi^{m} \| \\
& \leq \frac{(1-\tilde w)\delta_i + \sum_{j\neq i, j \in \mathcal{U}} \tilde w_j \delta_j} {\beta}((\eta\beta + 1)^m - 1).
\end{aligned}
\end{equation}
\label{lemma: local parameter difference}
\end{lemma}
\begin{IEEEproof}
Firstly, expand the left term with (\ref{equation: local update}) and \textbf{Definition \ref{definition: auxiliary parameter vector}}. Then, the triangle inequality can be utilized after adding a zero term $\nabla F_i(\bm \phi^{m-1}) - \nabla F_i(\bm \phi^{m-1})$. We have the norm of the model difference as
\begin{equation}
\begin{aligned}
& \| \boldsymbol\theta_i^{m} - \boldsymbol\phi^{m} \| \\
& = \left\| \boldsymbol\theta_i^{m-1} - \eta \nabla F_i(\boldsymbol\theta_i^{m-1}) - (\boldsymbol\phi^{m-1} - \eta \nabla \tilde F(\boldsymbol\phi^{m-1})) \right\| \\
& \leq \eta \| \nabla \tilde F(\boldsymbol\phi^{m-1}) - \nabla F_i(\boldsymbol\phi^{m-1}) \| + \| \boldsymbol\theta_i^{m-1} - \boldsymbol\phi^{m-1} \| \\
& \quad + \eta \| \nabla F_i(\boldsymbol\theta_i^{m-1}) - \nabla F_i(\boldsymbol\phi^{m-1}) \|.
\end{aligned}
\label{equation: proof lemma1 1}
\end{equation}
In (\ref{equation: proof lemma1 1}), there are 3 norms of difference. With \textbf{Assumption \ref{assumption: smooth}}, the third term is simplified into
$
\eta \| \nabla F_i(\boldsymbol\theta_i^{m-1}) - \nabla F_i(\boldsymbol\phi^{m-1}) \| \leq \eta \beta \| \boldsymbol\theta_i^{m-1} - \boldsymbol\phi^{m-1} \|.
$
However, the first term is difficult to simplify directly due to $\nabla \tilde F(\boldsymbol\phi^{m-1})$. We expand the first term and construct the specific form, that is
\begin{align}
 & \| {\nabla \tilde F\left( {\boldsymbol\phi^{m - 1} } \right) - \nabla F_i \left( {\boldsymbol\phi^{m - 1} } \right)} \| \notag \\
 & \leq \| {\left( {\tilde w_i  - 1} \right) \left(\nabla F_i \left( {\boldsymbol\phi^{m - 1} } \right) - \nabla F\left( {\boldsymbol\phi^{m - 1} } \right)\right)} \| \label{equation: proof lemma1 3}
 \\ &  \quad + \| {\sum\limits_{j \ne i} {\tilde w_j \nabla F_j \left( {\boldsymbol\phi^{m - 1} } \right)}  - \sum_{j \ne i} {\tilde w_j \nabla F\left( {\boldsymbol\phi^{m - 1} } \right)} } \|. \notag
\end{align}
Now \textbf{Assumption \ref{assumption: divergence}} can be used in (\ref{equation: proof lemma1 3}) and with the triangle inequality it is simplified into
\begin{equation}
\| \nabla \tilde F(\boldsymbol\phi^{m-1}) - \nabla F_i(\boldsymbol\phi^{m-1}) \| \leq (1- \tilde w_i) \delta_i + \sum_{j \neq i} \tilde w_j \delta_j = C_i.
\label{equation: proof lemma1 4}
\end{equation}
Substituting (\ref{equation: proof lemma1 4}) into (\ref{equation: proof lemma1 1}), a form available for recurrence is $ \| \boldsymbol\theta_i^{m} - \boldsymbol\phi^{m} \| + \frac{C_i}{\beta} \leq (\eta\beta + 1) (\| \boldsymbol\theta_i^{m-1} - \boldsymbol\phi^{m-1} \| + \frac{C_i}{\beta} )$. With the geometric progression from 0 to $m$, we have
\begin{equation}
\begin{aligned}
\| \boldsymbol\theta_i^{m} - \boldsymbol\phi^{m} \| &\leq \frac{C_i}{\beta} ((\eta\beta + 1)^m - 1) \\
& = \frac{(1- \tilde w_i) \delta_i + \sum_{j \neq i} \tilde w_j \delta_j}{\beta} ((\eta\beta + 1)^m -1).
\end{aligned}
\end{equation}
\end{IEEEproof}
Now, the difference norm of client $i$ is bounded, and the global difference norm can be expanded by
$
\| \boldsymbol\theta^{m} - \boldsymbol\phi^{m} \| \leq \| \boldsymbol\theta^{m-1} - \boldsymbol\phi^{m-1} \| + \eta \sum^U_{i=1} \tilde w_i \| \nabla F_i(\boldsymbol\theta_i^{m-1}) - \nabla F_i(\boldsymbol\phi^{m-1}) \|.
$
In addition, with \textbf{Assumption \ref{assumption: smooth}} and \textbf{Lemma \ref{lemma: local parameter difference}}, the second term can be further simplified into
\begin{equation}
\| \boldsymbol\theta^{m} - \boldsymbol\phi^{m} \| \leq  \| \boldsymbol\theta^{m-1} - \boldsymbol\phi^{m-1} \| + \eta((\eta\beta + 1)^{m-1} - 1) A_1,
\label{equation: proof theorem1 2}
\end{equation}
where $A_1 = 2\sum^U_{i=1}(\tilde w_i - \tilde w_i^2) \delta_i$. By accumulating two slides of (\ref{equation: proof theorem1 2}) from $m$ to 0, we have
\begin{equation}
\begin{aligned}
\| \boldsymbol\theta^{m} - \boldsymbol\phi^{m} \| & \leq  \frac{A_1}{\beta}((\eta\beta+1)^m - 1) - \eta m A_1 \\
& = \frac{A_1}{\beta} ((\eta\beta + 1)^m - \eta\beta m - 1).
\end{aligned}
\end{equation}
This completes the proof of \textbf{Theorem \ref{theorem: parameter difference}}.
\subsection{Proof of Theorem \ref{theorem: loss of auxiliary parameter}\label{subsection: proof of theorem 2}}
To prove \textbf{Theorem \ref{theorem: loss of auxiliary parameter}}, \textbf{Lemma \ref{lemma: bound of auxiliary parameter function difference}} is proposed as follows.
\begin{lemma}
The difference between functions of adjacent auxiliary parameters is bounded by
\begin{equation}
\begin{aligned}
& F(\boldsymbol\phi^{m+1}) - F(\boldsymbol\phi^{m}) \\
& \leq (\eta - \eta^2\beta) \sqrt{2\beta A_2 (F(\boldsymbol\phi^{0}) - F(\boldsymbol\theta^*))} + \frac{\eta^2 \beta A_2}{2} \\
& \quad - \frac{(2\eta - \eta^2\beta) (F(\boldsymbol\phi^{m}) - F(\boldsymbol\theta^*))^2}{2B_1^2},
\end{aligned}
\end{equation}
where $A_2 = 2\sum^U_{i=1}(\tilde w_i + w_i - 2 a_i w_i)\delta_i^2$ and $B_1 = \max_{m} \| \boldsymbol\phi^{m} - \boldsymbol\theta^* \|$.
\label{lemma: bound of auxiliary parameter function difference}
\end{lemma}
\begin{IEEEproof}
With \textbf{Definition \ref{definition: auxiliary parameter vector}} and \textbf{Lemma \ref{lemma: global loss function property}}, the auxiliary parameter function is expanded into
\begin{equation}
\begin{aligned}
& F(\boldsymbol\phi^{m+1}) - F(\boldsymbol\phi^{m}) \\
& \leq \nabla F(\boldsymbol\phi^{m})^{\rm T} (-\eta \nabla \tilde F(\boldsymbol\phi^ {m})) + \frac{\beta}{2} \| -\eta\nabla \tilde F(\boldsymbol\phi^{m}) \|^2.
\end{aligned}
\label{equation: proof lemma2 1}
\end{equation}
In (\ref{equation: proof lemma2 1}), $\tilde F(\boldsymbol\phi^{m})$ is hard to bound. Thus, a zero term $F(\boldsymbol\phi^{m}) - F(\boldsymbol\phi^{m})$ is added and $G$ is used to denote $\nabla F(\bm \phi^m)$ for the simplification. The norm of (\ref{equation: proof lemma2 1}) is split and (\ref{equation: proof lemma2 1}) is rearranged into
\begin{equation}
\begin{aligned}
& F(\boldsymbol\phi^{m+1}) - F(\boldsymbol\phi^{m}) \\
& \leq (\eta^2\beta - \eta) G^{\rm T} (\tilde G - G) + \frac{\eta^2\beta}{2} \| \tilde G - G \|^2 \\
& \quad + (\frac{\eta^2\beta}{2} - \eta) \| G \|^2.
\end{aligned}
\label{equation: proof lemma2 2}
\end{equation}
3 terms in the right side of (\ref{equation: proof lemma2 2}) are named by the cross term, the difference norm term and the norm term, respectively. With the Cauchy-Schwarz inequality, the cross term is bounded by
$
(\eta^2\beta - \eta) G^{\rm T} (\tilde G - G) \leq (\eta - \eta^2\beta) \| G \| \| \tilde G - G\|,
$
where the learning rate must satisfy $\eta < \frac{1}{\beta}$. For the difference norm, a zero term $\sum_{i \in \mathcal{U}_{\rm out}} w_i G - \sum_{i\in \mathcal{U}_{\rm in}} (\tilde w_i - w_i) G$ is added to get
\begin{equation}
\begin{aligned}
\| \tilde G - G \|^2 = & \Big\| \sum_{i \in \mathcal{U}_{\rm in}} (\tilde w_i - w_i) G_i - \sum_{i \in \mathcal{U}_{\rm in}} (\tilde w_i - w_i) G \\
& \enspace  + \sum_{i \in \mathcal{U}_{\rm out}} w_i G - \sum_{i \in \mathcal{U}_{\rm out}} w_i  G_i \Big\|^2.
\end{aligned}
\label{equation: proof lemma2 4}
\end{equation}
Since the square mean is greater than the arithmetic mean and $\|\cdot\|^2$ is convex, (\ref{equation: proof lemma2 4}) is split into
$
\| \tilde G - G \|^2 \leq 2(1 - \sum^U_{j=1} a_jw_j) (\sum_{i \in \mathcal U_{\rm out}} w_i \| G - G_i \|^2 + \sum_{i \in \mathcal U_{\rm in}} (\tilde w_i - w_i) \| G_i - G \|^2).
$
After splitting the norm, \textbf{Assumption \ref{assumption: divergence}} can be used to simplify the form into
$
\| \nabla \tilde F(\boldsymbol\phi^{m}) - \nabla F(\boldsymbol\phi^{m}) \|^2 \leq 2(1 - \sum^U_{j=1} a_jw_j)\sum^U_{i=1} (\tilde w_i + w_i - 2a_iw_i) \delta_i^2 = A_2.
$
So far, the cross term and difference norm term have already been transformed into closed form with acquired parameters. However, $\| \nabla F(\boldsymbol\phi^{m}) \|$ has not been bounded yet. To this end, \textbf{Lemma \ref{lemma: global loss function property}} is utilized to give an inequality as
\begin{equation}
\begin{aligned}
& F(\boldsymbol\phi^{m}) - F(\boldsymbol\theta^*) \\
& \geq \nabla F(\boldsymbol\theta^*)^{\rm T}(\boldsymbol\phi^{m} - \boldsymbol\theta^*) + \frac{1}{2\beta} \| \nabla F(\boldsymbol\phi^{m}) - \nabla F(\boldsymbol\theta^*) \|^2.
\end{aligned}
\label{equation: proof lemma2 7}
\end{equation}
By substituting $\nabla F(\boldsymbol\theta^*) = 0$ in (\ref{equation: proof lemma2 7}), the upper bound on the gradient norm is
$
\| \nabla F(\boldsymbol\phi^{m}) \| \leq \sqrt{2\beta(F(\boldsymbol\phi^{m}) - F(\boldsymbol\theta^*))}.
$
Except the upper bound, $\| \nabla F(\boldsymbol\phi^{m}) \|$ also needs a lower bound to enlarge $(\frac{\eta^2\beta}{2} - \eta) \| \nabla F(\boldsymbol\phi^{m}) \|^2$ (since the sign is negative). The convexity of $F(\boldsymbol\theta)$ gives $F(\boldsymbol\theta^*) \geq F(\boldsymbol\phi^{m}) + \nabla F(\boldsymbol\phi^{m})^{\rm T} (\boldsymbol\theta^* - \boldsymbol\phi^{m})$. And according to Cauchy-Schwarz inequality, the upper bound is given by
$
\| \nabla F(\boldsymbol\phi^{m}) \| \geq \frac{F(\boldsymbol\phi^{m}) - F(\boldsymbol\theta^*)}{\| \boldsymbol\phi^{m} - \boldsymbol\theta^* \|} \geq \frac{F(\boldsymbol\phi^{m}) - F(\boldsymbol\theta^*)}{B_1}.
$
After above derivations, we can further analyze (\ref{equation: proof lemma2 2}). Substituting derived upper bounds of 3 terms into (\ref{equation: proof lemma2 2}), an upper bound is obtained as
\begin{equation}
\begin{aligned}
& F(\boldsymbol\phi^{m+1}) - F(\boldsymbol\phi^{m}) \\
& \leq (\eta - \eta^2\beta) \sqrt{2\beta A_2(F(\boldsymbol\phi^{0}) - F(\boldsymbol\theta^*))} + \frac{\eta^2\beta A_2}{2}  \\
& \quad - \frac{(2\eta - \eta^2\beta) (F(\boldsymbol\phi^{m}) - F(\boldsymbol\theta^*))^2}{2B_1^2}.
\end{aligned}
\label{equation: proof lemma2 11}
\end{equation}
Note that the right side of (\ref{equation: proof lemma2 11}) should be negative. Specifically, if all clients participate, $A_2$ will vanish to 0. Then, the learning rate is turned down to guarantee $F(\boldsymbol\phi^{m})$ decreases as $m$, i.e., $F(\boldsymbol\phi^{\tau}) \leq \cdots \leq F(\boldsymbol\phi^{1}) \leq F(\boldsymbol\phi^{0})$. Enlarging the first term of the right side of (\ref{equation: proof lemma2 11}) with the decreasing property, we prove \textbf{Lemma \ref{lemma: bound of auxiliary parameter function difference}}.
\end{IEEEproof}
We are now ready to prove \textbf{Theorem \ref{theorem: loss of auxiliary parameter}}. The difference between loss functions of the auxiliary parameter and the optimal parameter can be written as a recursive formula. Firstly, (\ref{equation: proof lemma2 11}) subtracts $F(\boldsymbol\theta^*)$ in both sides and is rearranged into
\begin{equation}
\begin{aligned}
& F(\boldsymbol\phi^{m+1}) - F(\boldsymbol\theta^*) \\
& \leq F(\boldsymbol\phi^{m}) - F(\boldsymbol\theta^*) - \frac{(2\eta - \eta^2 \beta)(F(\boldsymbol\phi^{m}) - F(\boldsymbol\theta^*))^2}{2B_1^2} \\
& \quad + (\eta - \eta^2\beta) \sqrt{2\beta A_2(F(\boldsymbol\phi^{0}) - F(\boldsymbol\theta^*))} + \frac{\eta^2\beta A_2}{2}.
\end{aligned}
\end{equation}
To express it more concisely, $d_m$ is defined by $d_m = F(\boldsymbol\phi^{m}) - F(\boldsymbol\theta^*)$. Hence, dividing $d_m d_{m+1}$ in both sides, we have
\begin{align}
& \frac{1}{d_m} \leq \frac{1}{d_{m+1}} - \frac{(2\eta - \eta^2\beta)d_m^2} {2B_1^2 d_m d_{m+1}} \\
& \qquad + \frac{(\eta - \eta^2\beta) \sqrt{2\beta A_2(F(\boldsymbol\phi^{0}) - F(\boldsymbol\theta^*))}}{d_m d_{m+1}} + \frac{\eta^2\beta A_2}{2d_m d_{m+1}}.\notag
\end{align}
By enlarging the fraction according to $d_m$ decreasing property, it is rearranged into
\begin{equation}
\begin{aligned}
& \frac{1}{d_{m+1}} \geq \frac{1}{d_m} + \frac{2\eta - \eta^2\beta}{2B_1^2} \\
& \quad - \frac{(\eta - \eta^2\beta) \sqrt{2\beta A_2(F(\boldsymbol\phi^{0}) - F(\boldsymbol\theta^*))} + \frac{\eta^2\beta A_2}{2}}{d_m^2}.
\end{aligned}
\label{equation: proof theorem2 3}
\end{equation}
Accumulating (\ref{equation: proof theorem2 3}) from $m=0$ to $m = \tau -1$ and minifying $-\frac{1}{d_m}$ into $-\frac{1}{d_\tau}$, the relation between $d_\tau$ and $d_0$ is
$
\frac{1}{d_\tau} \geq \frac{1}{d_0} + \frac{(2\eta - \eta^2\beta)\tau}{2B_1^2} - \frac{\tau A_3}{d_\tau^2},
$
where $A_3 = ((\eta - \eta^2\beta)\sqrt{2\beta A_2 (F(\boldsymbol\phi^{0}) - F(\boldsymbol\theta^*))} + \frac{\eta^2\beta A_2}{2})$. Thus, we have a quadratic equation of $\frac{1}{d_\tau}$. Due to $d_\tau > 0$, the feasible region is $\frac{1}{d_\tau} \geq \frac{-1 + \sqrt{1 + \frac{4\tau A_3}{d_0} + \frac{(4\eta - 2\eta^2\beta)\tau^2 A_3}{B_1^2}}}{2\tau A_3}$. By substituting the definition of $d_m=F(\boldsymbol\phi^{m}) - F(\boldsymbol\theta^*)$, we have
\begin{equation}
\begin{aligned}
& F(\boldsymbol\phi^{\tau}) - F(\boldsymbol\theta^*) \\
& \leq \frac{2\tau A_3}{-1 + \sqrt{1 + \frac{4\tau A_3}{F(\boldsymbol\phi^{0}) - F(\boldsymbol\theta^*)} + \frac{(4\eta - 2\eta^2\beta)\tau^2 A_3}{B_1^2}}}.
\end{aligned}
\label{equation: proof theorem2 6}
\end{equation}
This completes the proof of \textbf{Theorem \ref{theorem: loss of auxiliary parameter}}.

\ifCLASSOPTIONcaptionsoff
  \newpage
\fi

\bibliographystyle{IEEEtran}
\bibliography{summary}

\end{document}